%% file: main.tex
\documentclass[a4paper,UKenglish,cleveref, autoref, thm-restate]{jsat}

\pdfoutput=1 %uncomment to ensure pdflatex processing (mandatatory e.g. to submit to arXiv)
\usepackage{algorithm}
\usepackage[noend]{algpseudocode}
\usepackage{tikz}
\usetikzlibrary{positioning, shapes.geometric, arrows.meta}
\usepackage{booktabs}

\def\l{{\ell}}

\def\var{{\texttt{var}}}
\def\score{{\texttt{score}}}

\newcommand{\n}[1]{\bar{#1}}

\newcommand\dsat{\textproc{\texttt{Dsat}}}
\newcommand\decide{\textproc{\textsc{Decide}}} 
\newcommand\undecide{\textproc{\textsc{Undecide}}} 
\newcommand\assert{\textproc{\textsc{Assert}}}

\newcommand\implied{\textproc{\textsc{Implied?}}} 
\newcommand\falsified{\textproc{\textsc{Falsified?}}}

\newcommand\acl{\textproc{\textsc{Get-Asserting-Clause}}}

\newcommand\addclause{\textproc{\textsc{Add-Asserting-Clause}}}

\newcommand\pafter{\textproc{\textsc{Pruned-After}}}
\newcommand\getl{\textproc{\textsc{Get-Literal}}}
\newcommand\getc{\textproc{\textsc{Get-Clause}}}

\newcommand\selectl{\textproc{\textsc{Select-Literal}}}
\newcommand\updatescore{\textproc{\textsc{Update-Score}}}
\newcommand\cnfstate{\textproc{\textsc{CNF-State}}}
\newcommand\onestate{\textproc{\textsc{Fixed-State?}}}
\newcommand\onebit{\textproc{\textsc{On-Bit}}}
\newcommand\allbits{\textproc{\textsc{On-Bits}}}
\newcommand\singlebit{\textproc{\textsc{Single-On-Bit?}}}
\newcommand\astate{\textproc{\textsc{Active-State}}}
\newcommand\lastpstate{\textproc{\textsc{Last-Pruned-State}}}
\newcommand\analyzel{\textproc{\textsc{Disjoin-Literal}}}

\newcommand\true{\texttt{true}}
\newcommand\false{\texttt{false}}
\newcommand\nil{\texttt{nil}}

\newcommand\ST{{\Pi}} % cnf state
\newcommand\vs{\text{\em s}} % state
\newcommand\dlevel{\texttt{dLevel}} % decision level
\newcommand\plevel{\texttt{pLevel}} % pruning level
\newcommand\astates{\texttt{aStates}} % active states
\newcommand\ptime{\texttt{pTime}} % pruning time
\newcommand\pvars{\texttt{pVars}} % saved active states
\newcommand\watchedby{\texttt{watchedBy}}

\newcommand\lstack{\texttt{lStack}} % decision-level stack
\newcommand\vstack{\texttt{vStack}} % variable stack
\newcommand\tstack{\texttt{tStack}} % timer stack
\newcommand\reason{\texttt{reason}} % reason for pruning
\newcommand\tops{\texttt{top}} % top of stack
\newcommand\cclause{\texttt{cClause}} % conflict clause

\newcommand\card{\texttt{card}}
\newcommand\idx{\texttt{index}}
\newcommand\vcount{\texttt{vCount}}
\newcommand\clauses{\texttt{clauses}}
\newcommand\uclauses{\texttt{uClauses}}
\newcommand\literals{\texttt{literals}} 
\newcommand\literal{\texttt{literal}} 
\newcommand\states{\texttt{states}}
\newcommand\state{\texttt{state}}

\newcommand\cstates{\texttt{cStates}}
\newcommand\lastcstate{\texttt{lastcState}}
\newcommand\nasserting{\texttt{aCount}}

\newcommand{\lpair}[2]{\langle #1, #2 \rangle}

\newcommand\bump{\texttt{bump}}
\newcommand\bumpInc{\texttt{bumpInc}}
\newcommand\lclause{\mathit{lclause}}
\newcommand\success{\mathit{success}}

\renewcommand\lnot[1]{\overline{#1}}
\renewcommand\land{\cdot}
\renewcommand\lor{+}

\newcommand\bAND{\,\&\,}

\newcommand\NOT[1]{\overline{#1}}
\newcommand\AND{\cdot}
\newcommand\OR{+}

\newcommand\w{\omega}

\newcommand{\inst}{{\cal I}}

\newcommand{\refAppCode}{Appendix~\ref{app:pcode}}

\title{\texttt{Dsat}: A Native SAT Solver for Discrete Logic}

\author{Yaofang Zhang}{Department of Computer Science, University of California, Los Angeles}{vzhang@cs.ucla.edu}{https://orcid.org/0009-0001-2686-8954}{}
\author{Ken Zhou}{Department of Computer Science, University of California, Los Angeles}{kbzhou01@ucla.edu}{https://orcid.org/0009-0002-6206-7932}{}

\author{Adnan Darwiche}{Department of Computer Science, University of California, Los Angeles}{darwiche@cs.ucla.edu}{https://orcid.org/0000-0003-3976-6735}{}

\authorrunning{Y. Zhang, K. Zhou, A. Darwiche} %TODO mandatory. First: Use abbreviated first/middle names. Second (only in severe cases): Use first author plus 'et al.'

\Copyright{Yaofang Zhang, Ken Zhou, Adnan Darwiche} %TODO mandatory, please use full first names. LIPIcs license is "CC-BY";  http://creativecommons.org/licenses/by/3.0/


 %TODO mandatory; please add comma-separated list of keywords

\supplementdetails{Software}{https://github.com/uclareasoning/dsat} %linktext, cite, and subcategory are optional

\funding{This work has been partially supported by ONR grant \#N000142612296.}
\nolinenumbers%uncomment to disable line numbering

\EventNoEds{2}
\EventLongTitle{This article is submitted to JSAT}
\EventShortTitle{JSAT YYYY}
\EventAcronym{JSAT}
\EventYear{YYYY}
\EventDate{}
\EventLocation{Little Whinging, United Kingdom}
\EventLogo{}
\SeriesVolume{VV(I))} % Need to match!
\ArticleNo{XX(Y)} % Need to match!
\Submitted{MMMM YYYY}
\Revised{MMMM YYYY}
\begin{document}
\maketitle

\begin{abstract}
Discrete variables are common in many applications, such as probabilistic reasoning, planning and explainable AI.
When symbolic reasoning techniques are brought in to bear on these applications, a standard technique for handling discrete variables is to binarize them into Boolean variables to allow the use of Boolean computational machinery such as SAT solvers. 
This technique can face both computational and semantical challenges though. 
In this work, we develop a native SAT solver for discrete logic, which is a direct extension of Boolean logic in which variables can take arbitrary values. 
Our proposed solver has a similar design to Boolean SAT solvers, with ingredients such as unit resolution and clause learning but ones that operate natively on discrete variables. 
We illustrate the merits of the developed SAT solver by comparing it empirically to CSP solvers applied to discrete CNFs, to Boolean SAT solver applied to binarized CNFs, and to some hybrid solvers. 
\end{abstract}

\section{Introduction}

Boolean logic and its associated computational 
machinery have found much use in AI
and computer science over the last few decades,
including to problems that are 
not symbolic in nature~\cite{acm/fichte2023silent}. 
For example, beyond symbolic applications
like verification, synthesis and planning, Boolean logic played
an influential role in probabilistic reasoning~\cite{kr/Darwiche02,ai/ChaviraD08}
and more recently explainable AI; see, e.g.,~\cite{lics/Darwiche23,ijcai/ShihCD18,kr/AudemardKM20,kr/AudemardBBKLM21,IgnatievNM19a,aiia/IgnatievNA020}. 
Many of the applications targeted by Boolean logic
involve discrete or discretized variables that can have many values, which are handled using binarization techniques.
%where a discrete variable with $n$ values may be encoded using $n$ Boolean variables; see~\cite{corr/abs-2007-01493} for a review of several encoding schemes.
The treatment of discrete variables through binarization,
instead of natively, is usually motivated by the potent computational machinery for Boolean logic, particularly SAT solvers. 
This can be seen, for example, in the constraint satisfaction (CP) literature where in recent CSP solver competitions,\footnote{MiniZinc Challenge~\cite{DBLP:journals/aim/StuckeyFSTF14} (also \url{https://www.minizinc.org/challenge}) and XCSP3 Competition~\cite{DBLP:journals/corr/abs-2312-05877}.} the leading CSP solvers --- PicatSAT~\cite{DBLP:series/sbis/ZhouKF15,DBLP:conf/padl/ZhouK16} and OR-Tools CP-SAT~\cite{cpsatlp,DBLP:conf/cp/PerronDG23} --- both rely on binarization techniques that represent CSPs using Boolean variables.\footnote{A central technique used by CP-SAT
is lazy clause generation (LCG)~\cite{DBLP:conf/cpaior/Stuckey10}, which incrementally (i.e., lazily) generates Boolean variables and Boolean clauses that represent constraints on the original (discrete) variables of the problem. CP-SAT also uses standard SAT technology of clause learning. This solver is developed by Google as part of their OR-Tools library (\url{https://developers.google.com/optimization/cp/cp_solver}).
PicatSAT focuses more on optimizing the encoding of discrete variables into Boolean variables, e.g., using techniques in~\cite{DBLP:conf/cp/ZhouK17}. 
PicatSAT then calls an existing state-of-the-art SAT solver, e.g., kissat~\cite{BiereFleury-SAT-Competition-2022-solvers}, on the transformed problem.}

The binarization of discrete
variables can pose both computational and semantical challenges though, 
calling for additional efforts that treat discrete variables natively. 
In explainable AI, for example, it has been argued that such native treatment 
can be critical for the proper formalization of 
certain explanations~\cite{jelia/JiD23,lics/Darwiche23}.
In this work, we take a further step in this direction by proposing a 
conflict-driven clause-learning (CDCL) SAT solver
for \textit{discrete} CNFs, also called \textit{signed} 
CNFs in~\cite{signedSAT2000}. 
A discrete CNF is a conjunction of discrete clauses,
where a discrete clause is a disjunction of discrete literals.
A discrete literal is a subset of the values for a discrete variable.
Hence, discrete CNFs provide a native (direct) representation of clausal 
constraints over discrete variable that does not rely on 
binarization.\footnote{The lifting of Boolean logic syntax and semantics to a setting based on discrete 
variables has been pursued under various names in the past, including 
many-valued logic, signed logic, multiple-valued logic, discrete logic, etc.
We prefer the term \textit{discrete logic}
as the other terms may give the impression that we are lifting the notion of a truth value (which has also been proposed in earlier AI works).} 

Discrete CNFs were initially studied in~\cite{signedSAT2000}, 
where it was shown that their satisfiability remains NP-complete for binary clauses. Additional results included a corresponding resolution rule and a DPLL algorithm for satisfiability. Later work~\cite{sat/csp2dsat2007}
studied the connections between the closure of discrete CNFs under (unit) resolution and 
the application of various notions of consistency to the corresponding CSPs.
As far as we know, there has been only two prior works~\cite{iccad/CAMA2003,cp/mvsat2010} on developing native SAT solvers for discrete 
CNFs but they do not seem to have provided a practical alternative to binarization approaches (more on this later).
The need for ``a highly efficient many-valued SAT solver incorporating the recently developed SAT technology'' was later emphasized in~\cite{ansotegui2015sat}, which 
is the task we undertake in this work.
We will describe our solver using detailed pseudocode  
in the spirit of works such as~\cite{minisat_2004}.

An earlier version of this work appeared in the proceedings of the 29th International Conference on Theory and Applications of Satisfiability Testing (SAT 2026)~\cite{dsatSAT2026}.
The article extends the conference version with additional theoretical and experimental results, as well as an appendix providing more detailed descriptions of the data structures and pseudocode in our system.

This paper is structured as follows. 
We start in \cref{sec:dlogic} by defining the syntax and semantics
of discrete logic, where we motivate it further from the lens of explainable AI.
\cref{sec:ur} discusses unit resolution in discrete logic and 
proposes a lazy algorithm similar to what is currently employed by Boolean SAT solvers. 
\cref{sec:cdcl} presents our proposed CDCL SAT solver for discrete CNFs, called
\dsat, after
proposing a clause learning algorithm based on unit resolution proofs, and 
a method for making decisions based on discrete literals.
\cref{sec:experiments} provides an empirical evaluation of \dsat, based on a number
of benchmark families, and against a number of baseline solvers.
\cref{sec:conclusion} closes with some concluding remarks. 

\section{Discrete Logic}\label{sec:dlogic}

We start with the syntax, semantics and notational conventions 
of discrete logic as in~\cite{lics/Darwiche23,aaai/DarwicheJ22}.

{\bf Syntax.} 
Each discrete variable \(X\) has a finite number of 
\textit{states} \(x_1, \ldots, x_n,\) \(n > 1\). 
A \textit{literal} $\l$ for variable \(X\), called an \(X\)-literal,
is a set of states such that \(\emptyset \subset \l \subset \{x_1, \ldots, x_n\}\). 
We often denote a literal such as \(\{x_1,x_3,x_4\}\) 
by \(x_{134}\) which reads: the state of variable \(X\) 
is either \(x_1\) or \(x_3\) or \(x_4\).
A literal is \textit{simple} iff it contains
a single state. Hence, \(x_3\) is a simple literal but
\(x_{134}\) is not.
If a variable \(X\) has two states, it is said to be \textit{Boolean} or  \textit{binary}
and its states are denoted by $x$ and $\NOT x$.
Such a variable has only two literals
\(\{x\}\) and \(\{\NOT{x}\}\), which are simple and
denoted by \(x\) and \(\NOT x\) for convenience. 
A \textit{formula} is either a constant \(\top\), \(\bot\),
literal \(\l\), negation \(\NOT \alpha\), conjunction
\(\alpha \AND \beta\) or disjunction \(\alpha \OR \beta\) 
where \(\alpha\), \(\beta\) are formulas. 
%We will sometimes say ``discrete formula'' instead of ``formula'' when discussing Boolean formulas in the same context.
%We find it convenient to treat literals as both sets and as formulas. 
%Hence, for literals $\l_1$ and $\l_2$,
%$\l_1 \cap \l_2$ and $\l_1 \AND \l_2$ are exchangeable.

{\bf Semantics.} The semantics of discrete logic follow directly from
the semantics of Boolean logic. A \textit{world,} 
denoted \(\w\), maps 
each variable $X$ in \(\Sigma\) to one of its states, 
denoted $\w[X]$. A discrete formula $\alpha$ can 
be turned into a Boolean formula 
under the variable settings of world $\w$ as follows.
We replace each
$X$-literal $\l$ in $\alpha$ with $\top$ if $\w[X] \in \l$
and with $\bot$ otherwise, which leads to a Boolean formula 
$\alpha_\w$ that has no variables.
We say that world $\w$ \textit{satisfies} discrete 
formula $\alpha$ iff the corresponding Boolean formula 
$\alpha_\w$ evaluates to $\top.$ 
We also say in this case that world $\w$
is a \textit{model} of $\alpha.$ 
Formula \(\alpha\) implies/entails formula \(\beta\), 
written \(\alpha \models \beta\), iff
every model of \(\alpha\) is also a model of \(\beta\).
In this case, we say  $\alpha$ is \textit{stronger} than $\beta,$ and that $\beta$ is \textit{weaker} than $\alpha$.
A formula is \textit{satisfiable} (\textit{consistent}) iff it has at least one model.
A formula is \textit{valid} iff it is satisfied by
every world. 
The constant \(\top\) will also be used to denote 
a valid formula, and \(\bot\) will be used to denote an
unsatisfiable one. 

{\bf Normal Forms.} A \textit{term} is a conjunction of literals for distinct variables. 
A \textit{clause} is a disjunction of literals for distinct variables.
A term/clause is \textit{simple} if all its literals are simple.
Terms and clauses are often represented as sets (of their literals) so an empty term is valid and an empty clause is
unsatisfiable.
A \textit{Disjunctive Normal Form (DNF)} is a disjunction of terms.
A \textit{Conjunctive Normal Form (CNF)} is a conjunction of clauses.
A \textit{Negation Normal Form (NNF)} is a formula without negations.
%DNFs and CNFs are often represented as sets (of their terms/clauses) so an empty DNF is unsatisfiable and an empty CNF is valid.

{\bf Explainable AI.} 
The need to treat discrete variables natively, at least semantically,
arises when explaining the decisions
of classifiers; see~\cite{lics/Darwiche23}. 
In this context, a \textit{feature}
corresponds to a discrete variable \(X\) and 
a \textit{characteristic} corresponds to one of 
its simple literals $\{x_i\}$.
An \textit{instance} corresponds to a conjunction of
simple literals (characteristics), one for each feature.
A classifier can then be represented using a set of 
\textit{class formulas,} which are discrete formulas in
one-to-one correspondence with the classes of the classifier.

Consider the following decision graph which has three ternary features \(X, Y, Z\) and three classes $c_1, c_2, c_3$.% wrapfig is not allowed
\begin{figure}[H]
\centering
\input{figures/class_formula.tex}
\end{figure}
\noindent This classifier has \(27\) instances, 
partitioned as follows:
\(20\) instances in class \(c_1\), \(3\) instances in class \(c_2\) and \(4\) instances in class \(c_3\). 
Classes $c_1,c_2,c_3$ can be represented using the
following discrete formulas:
$\Delta_1 = x_{12}\OR x_3 \AND y_1 \AND z_{13}$, 
$\Delta_2 = x_3 \AND z_2$ and
$\Delta_3 = x_3 \AND y_{23} \AND z_{13}$. 
For example,
instance $\inst = x_3 \AND y_2 \AND z_2$ is
in class \(c_2\) since $\inst \models \Delta_2$.
Representing classifiers using class formulas
allows one to reason about such classifiers 
independently of their forms which can vary significantly.

One of the more recent developments in explainable AI  
notes that earlier treatments of explanations,  
based on Boolean features, lead to suboptimal explanations when
lifted to a discrete setting in the obvious way. As a result, a more 
general treatment has been provided in the context of discrete logic, based on 
new logical operators for computing instance abstractions that are
native discrete~\cite{jelia/JiD23}.
These and similar developments put discrete logic into focus 
recently. For example, they highlighted the need for computing prime implicants and implicates of discrete formulas, a subject
that received almost no attention in the literature as discussed 
in~\cite{jelia/JiD23}; see 
also~\cite{corr/abs-2007-01493} for difficulties that
arise when attempting to perform this computation using binarization.
A native SAT solver for discrete logic can help in these and similar computations.
The compilation of class formulas into tractable forms can also benefit from such a solver (more on this in~\cref{sec:experiments}).

\section{Unit Resolution on Discrete CNFs}\label{sec:ur}

Unit resolution is an inference rule in Boolean logic which
plays a critical role in Boolean SAT solving. The rule
allows us to infer clause $\alpha$ from literal $\l$ and 
clause $\lnot \l \lor \alpha,$ and is used by SAT solvers
to simplify CNFs and to efficiently detect some contradictions
when searching for satisfying assignments.
We now discuss unit resolution in discrete logic as it will
also play a critical role in discrete SAT solving.
Our definition of this inference rule will be based
on the resolution rule for discrete clauses as defined in~\cite{jelia/JiD23},
a slight variation on the one in~\cite{signedSAT2000}.

\begin{definition}\label{def:resolution}
Let $\alpha = \l \lor \sigma_1$ and $\beta = 
\l' \lor \sigma_2$ be two clauses, where 
$\l$ and $\l'$ are $X$-literals, 
$\l \not \models \l'$ and
$\l' \not \models \l$.
The resolvent of $\alpha$ and $\beta$ over variable $X$ is $\sigma = (\l \land \l') \lor \sigma_1 \lor \sigma_2$.
\end{definition}
Requiring $\l \not \models \l'$ and
$\l' \not \models \l$ ensures that the resolvant $\sigma$ is not subsumed by either $\alpha$ or $\beta$. 
When variable $X$ is Boolean, this condition ensures that literals $\l$ and $\l'$
are conflicting.

Unit resolution arises when at least one of the resolved clauses is unit (has a single literal).
\begin{definition}\label{def:ur}
Let $\l_1$ and $\l_2$ be $X$-literals such that
$\l_1 \not \models \l_2$ and
$\l_2 \not \models \l_1$.
Resolving literal $\l_1$ with clause 
$\l_2 \lor \sigma,$ called a \underline{unit resolution,} gives the resolvent $(\l_1 \land \l_2) \lor \sigma.$
\end{definition}

If $\l_1 \land \l_2 = \bot$, which always holds when variable $X$ is binary, the result of unit resolution is the shorter clause $\sigma$ which does not mention variable $X$.
In discrete logic, unit resolution is defined even when $\l_1 \land \l_2 \neq \bot$. In this case, the resolvent $(\l_1 \land \l_2) \lor \sigma$ subsumes clause
$\l_2 \lor \sigma$ but is not necessarily shorter so 
variable $X$ is not eliminated from 
clause $\l_2 \lor \sigma$. 

\subsection{Inferential Power of Discrete Unit Resolution}
\label{app:ur-power}

As an inference rule,
discrete unit resolution is more complete than Boolean unit resolution when the latter is applied to certain binarizations of the given CNF.
This is the case, for example, if we use the standard binarization in which we:
\begin{itemize}
    \item represent a discrete variable $X$ with states $\{1, \ldots, k\}$ using binary variables $x_1, \ldots, x_k$
    \item represent a literal $x_S$ using the Boolean clause $\bigvee_{i \in S} x_i$
    \item add to the CNF the exactly-one constraints by combining the at-least-one constraints ($x_1 \lor \ldots \lor x_k$), and the pairwise encoding of the at-most-one constraints ($\lnot x_i \lor \lnot x_j$ for $i \neq j$)
\end{itemize}
For an example, suppose variables $X$ and $Y$ have states $\{1,2,3,4\}$ each and consider the discrete
CNF $\Delta = (x_{13} \lor y_{12}) \land (x_{24}).$
Discrete unit resolution allows us to infer the literal $y_{12}$
from CNF $\Delta,$ which tells us that states $3$ and $4$
of variable $Y$ are impossible. 
The standard binarization of $\Delta$ is the CNF $\Delta_b$ shown below:
\begin{eqnarray*} 
&
(x_1 \lor x_3 \lor y_1 \lor y_2) \land (x_2 \lor x_4) \land
(x_1 \lor x_2 \lor x_3 \lor x_4) \land 
(y_1 \lor y_2 \lor y_3 \lor y_4) \land \\
&
(\NOT{x}_1 \lor \NOT{x}_2) \land (\NOT{x}_1 \lor \NOT{x}_3) \land (\NOT{x}_1 \lor \NOT{x}_4) \land (\NOT{x}_2 \lor \NOT{x}_3)
\land (\NOT{x}_2 \lor \NOT{x}_4) \land (\NOT{x}_3 \lor \NOT{x}_4) \land \\
&
(\NOT{y}_1 \lor \NOT{y}_2) \land (\NOT{y}_1 \lor \NOT{y}_3) \land (\NOT{y}_1 \lor \NOT{y}_4) \land (\NOT{y}_2 \lor \NOT{y}_3)
\land (\NOT{x}_2 \lor \NOT{x}_4) \land (\NOT{x}_3 \lor \NOT{x}_4).
\end{eqnarray*}

This Boolean CNF $\Delta_b$ implies the 
clause $\alpha = y_1 \lor y_2$
but it does not contain any unit clauses so 
we cannot derive $\alpha$ from $\Delta_b$ using unit resolution.
Neither can we refute $\Delta_b \land \NOT{y}_1 \land \NOT{y}_2$
using unit resolution. Hence, we cannot show  
$\Delta_b \models y_1 \lor y_2$ either by derivation or refutation using Boolean unit resolution, yet we could derive
literal $y_{12}$ from $\Delta$ using discrete unit resolution.
Hence, discrete unit resolution is more complete than Boolean unit resolution in this case.

Besides the pairwise encoding, the sequential counters encoding~\cite{sequentialCounterEncoding} is another common technique for the at-most-one constraints.
To encode that at most one of binary variables $x_1, \ldots, x_k$ is true,
this method requires introducing auxiliary variables $s_1, \ldots, s_{k-1}$.
Each $s_i$ can be viewed as an indicator for whether any $x_j$ ($j \le i$) is true: this relation is encoded as $x_i \implies s_i$ and $s_i \implies s_{i+1}$.
With the auxiliary variables, the at-most-one semantics can then be encoded as $s_i \implies \lnot x_{i+1}$.
To binarize the discrete CNF above, we can add auxiliary variables $s_1, s_2, s_3, t_1, t_2, t_3$, and get the following Boolean CNF:
\begin{eqnarray*}
&
(x_1 \lor x_3 \lor y_1 \lor y_2) \land (x_2 \lor x_4) \land
(x_1 \lor x_2 \lor x_3 \lor x_4) \land 
(y_1 \lor y_2 \lor y_3 \lor y_4) \land \\
&
(\lnot x_1 \lor s_1) \land (\lnot x_2 \lor s_2) \land (\lnot x_3 \lor s_3) \land (\lnot x_2 \lor \lnot s_1) \land (\lnot x_3 \lor \lnot s_2) \land (\lnot x_4 \lor \lnot s_3) \land (\lnot s_1 \lor s_2) \land (\lnot s_2 \lor s_3) \land \\
&
(\lnot y_1 \lor t_1) \land (\lnot y_2 \lor t_2) \land (\lnot y_3 \lor t_3) \land  (\lnot y_2 \lor \lnot t_1) \land (\lnot y_3 \lor \lnot t_2) \land (\lnot y_4 \lor \lnot t_3) \land (\lnot t_1 \lor t_2) \land (\lnot t_2 \lor t_3).
\end{eqnarray*}
\noindent This Boolean CNF $\Delta'_b$ still does not contain any unit clauses, and neither can we refute $\Delta'_b \land \lnot y_1 \land \lnot y_2$ using unit resolution. 
Thus, for the two common binarization techniques discussed so far, discrete unit resolution has more inferential power than Boolean unit resolution.

A variant binarization was discussed in~\cite{corr/abs-2007-01493} in the context of using algorithms
for computing prime implicants of Boolean formulas
to compute prime implicants for discrete formulas. 
This variant is based on representing a discrete literal $x_S$ using the Boolean term $\bigwedge_{i \not \in S} \lnot x_i$
instead of the Boolean clause $\bigvee_{i \in S} x_i$
since the two are equivalent given exactly-one constraints. If we apply this technique to the discrete CNF $\Delta$ above, we get the formula
\begin{eqnarray*} 
&
([\NOT{x}_2 \land \NOT{x}_4] \lor [\NOT{y}_3 \land \NOT{y}_4]) \land 
(\NOT{x}_1) \land (\NOT{x}_3) \land
(x_1 \lor x_2 \lor x_3 \lor x_4) \land 
(y_1 \lor y_2 \lor y_3 \lor y_4) \land \\
&
(\NOT{x}_1 \lor \NOT{x}_2) \land (\NOT{x}_1 \lor \NOT{x}_3) \land (\NOT{x}_1 \lor \NOT{x}_4) \land (\NOT{x}_2 \lor \NOT{x}_3)
\land (\NOT{x}_2 \lor \NOT{x}_4) \land (\NOT{x}_3 \lor \NOT{x}_4) \land \\
&
(\NOT{y}_1 \lor \NOT{y}_2) \land (\NOT{y}_1 \lor \NOT{y}_3) \land (\NOT{y}_1 \lor \NOT{y}_4) \land (\NOT{y}_2 \lor \NOT{y}_3)
\land (\NOT{x}_2 \lor \NOT{x}_4) \land (\NOT{x}_3 \lor \NOT{x}_4),
\end{eqnarray*}
which contains the unit clauses $\NOT{x}_1$ and $\NOT{x}_3.$
The above formula is not a CNF though. But we can
unfold the component
$([\NOT{x}_2 \land \NOT{x}_4] \lor [\NOT{y}_3 \land \NOT{y}_4])$ into
$
(\NOT{x}_2 \lor \NOT{y}_3) \land 
(\NOT{x}_2 \lor \NOT{y}_4) \land 
(\NOT{x}_4 \lor \NOT{y}_3) \land 
(\NOT{x}_4 \lor \NOT{y}_4)
$
to obtain a Boolean CNF, $\Delta''_b.$ Unit resolution
will now trigger on CNF $\Delta''_b$ which allows us to
derive the clause $x_2 \lor x_4$ but not 
$y_1 \lor y_2$. We cannot refute 
$\Delta''_b \land \NOT{y}_1 \land \NOT{y}_2$ using 
unit resolution either.
Again, discrete unit resolution is more complete in this case.

Furthermore, 
the first and third binarization techniques discussed above lead to Boolean CNFs whose size is quadratic in the cardinalities of discrete variables; 
the second technique can add fewer clauses (linear with respect to the variable cardinalities), but requires introducing auxiliary variables.
Moreover, for the third binarization technique, a discrete clause of length $l$
leads to a number of Boolean clauses that is exponential in $l$. 
Despite the additional variables or the considerable increase in CNF sizes,
Boolean unit resolution is not as complete as its discrete counterpart.
This further illustrates the merit of treating discrete CNFs natively.

Another perspective to examine the effect of binarization is to compare the discrete and Boolean CNFs in terms of the strength of their clauses. 
We will next introduce a measurement of strength for discrete (and Boolean) clauses and utilize it later in our empirical studies.

\begin{definition}
The looseness of an $X$-literal $\l$ is $- \log_2(1 - \frac{L}{C})$, where $L$ is the number of states in literal $\l$, and $C$ is the cardinality of variable $X$. 
The looseness of a clause is the sum of its literals' looseness.
\label{def:looseness}
\end{definition}

\noindent For a variable $X$ with cardinality $C$, as an $X$-literal $\l$ becomes stronger, 
$\l$'s looseness decreases from $\log_2 C$ ($\l$ contains $C-1$ states) to $\log_2(\frac{C}{C-1})$ ($\l$ contains only one state).\footnote{Note that as $C$ increases, $\log_2(\frac{C}{C-1})$ decreases asymptotically toward zero.}
For a constraint (literal or clause) with looseness score $m$, the fraction of the search space forbidden by the constraint is $1/{2^m}$;
thus, if we assume every world (complete assignment) is equally likely, looseness $m$ is exactly the Shannon information of the constraint's violation.
A clause's looseness can be viewed as its effective clause length:
for a clause that mentions only Boolean variables, its looseness simply equals to its number of literals.
In the discrete CNF $\Delta = (x_{13} \lor y_{12}) \land (x_{24})$, the two clauses have looseness 2 and 1 respectively, and an average of 1.5.
For both of the binarization techniques discussed earlier, 
binarizing $\Delta$ would result in a Boolean CNF that has average looseness over 2.

\subsection{A Lazy Algorithm}\label{sec:ur-alg}
\label{sec:watching}

It is well known that a Boolean CNF can be closed under
unit resolution in time linear in the CNF size~\cite{UR_linear}. One can
easily implement a procedure for unit resolution on discrete CNFs that is
also linear in the CNF size, where the notion of size is defined as follows.

\begin{definition}\label{def:csize}
The size of a clause $\l_1 \lor \ldots \lor \l_n$ is $|\l_1| + \ldots + |\l_n|$, where $|\l_i|$ denotes the number of states in literal $\l_i$ (size of $\l_i$). 
The size of a CNF is the sum of its clauses' sizes.
\end{definition}
A binary variable $X$ has only two literals, 
$\{x\}$ and $\{\n{x}\}$, 
each containing one state. Hence, if all variables in a clause are binary, the clause size is its length according to Definition~\ref{def:csize}.

A key advance in Boolean SAT solving is the technique of \textit{two-literal watching} for implementing unit resolution~\cite{Chaff},
which provides a lazy way for detecting whether a clause has become unit; that is, whether all its literals but one have been falsified. This is important because a unit clause --- corresponds to a variable setting in Boolean logic --- triggers further unit resolutions. Two-literal watching is based on ``watching'' two literals in each clause $\beta$ and visiting that clause only if one of these literals $\l$ has been falsified (i.e., $\lnot{\l}$ has been implied). 
A number of cases may arise when clause $\beta$ is visited:
the clause is now unit (every other literal in $\beta$ but one is falsified);
the clause is now subsumed (some literal in $\beta$ is implied); or
the clause is now empty (all its literals have been falsified) so we have a contradiction. If none of these cases materialize,  
another literal in $\beta$ gets watched instead of literal $\l$.

We will next propose a similar lazy algorithm for unit resolution on
discrete CNFs, which will become a backbone of our proposed SAT solver.
The first observation is that in a Boolean setting,
a unit clause sets the state of a variable as it corresponds to 
either $X=\true$ or $X=\false$ for some Boolean variable $X$. 
However, in the discrete setting, a unit clause is a literal $\l$ which corresponds to some states of some discrete variable $X$. The discrete unit clause does not set the state of variable $X$, but it prunes (falsifies) some of its states: those outside literal $\l.$
This leads to the following interpretation of discrete unit resolution,
which will form the basis for our lazy algorithm. 
If we have the unit clause $\l,$ then all states $\vs$ in literal $\lnot{\l}$ are impossible and can be \textit{pruned} from the CNF. Pruning a state $\vs$ of some variable $X$ amounts to removing $\vs$ from all $X$-literals in the CNF. 
This pruning process strengthens clauses in the CNF and may lead to 
(1)~deriving a literal $\l'$ if all literals in a clause, except for literal $\l'$, lose all their states, or
(2)~establishing a contradiction if all literals in a clause lose their states. 
If a literal $\l'$ is derived,
we know that all states in $\lnot{\l'}$ are impossible, which means that more state pruning can be done. 
The process of state pruning continues until a contradiction is detected, or until
there are no more states to prune. 

\cref{alg:ur} provides our lazy implementation of unit resolution, which is based on the observation that 
not only do we need to watch two literals in each clause, but that
each literal that is being watched must also watch one of its states.
The reason is that not only do we need a lazy way for detecting whether
a clause has become unit (or empty) but we also need a lazy way for detecting
whether a literal has been falsified (i.e., all its states have been pruned).
The clauses watching a literal $\l$ are tracked in $\l.\watchedby$, and the literals watching a states $\vs$ are tracked in $\vs.\watchedby$.
We stress here that only literals that are being watched by clauses need to watch one of their states. 
Moreover, our technique does not require a literal to keep track of the state it is watching; only a state needs to keep track of the literals watching it. This is quite important for efficiency since a significant number of literals can get created 
during SAT solving as we shall see later; that is, literals that do not appear in the input CNF. This is a fundamental difference with Boolean CNFs where we have a fixed number of literals that are part of the input CNF and whose number is at most twice the number of Boolean variables (a discrete variable with $m$ values has $2^m-2$ literals). This difference will become more apparent when we discuss clause learning in \cref{sec:clearn}.

The three functions in \cref{alg:ur}
-- \decide($\l$,$\ST$), \undecide($\ST$), and \assert($\l$,$\ST$) --
constitute an interface to unit resolution that will be utilized by the
upcoming SAT solver. Here, 
$\ST$ is a CNF \textit{state,} which contains 
a CNF and some bookkeeping data. 
For example, $\ST.\dlevel$ is the \textit{decision level}, 
which is initially set to $0$
and incremented (decremented) every time we call \decide\ (\undecide).
\decide($\l$,$\ST$) adds the literal (unit clause) $\l$ to the CNF, where literal $\l$ is also called a \textit{decision} in this case. It then
runs unit resolution, and returns \true\ or \false, depending on whether unit resolution finds a contradiction.
If \false\ is returned, the clause that became empty is stored in
$\ST.\cclause$ and used later for learning a clause. 
Otherwise, the CNF state $\ST$ is changed into a new state $\ST'$ which can be rolled back using \undecide($\ST'$).
The functions \decide\ and \undecide\ do bookkeeping for the most part as the main implementation of unit resolution is embedded in the function \assert($\l$,$\ST$). 
The pair $\lpair{V}{I}$ in \cref{alg:ur} represents a literal since $V$ is a variable and $I$ are some states of variable $V$. 
Our implementation represents states $I$ using an array of integers so we can use bit-level operations when manipulating literals; see \refAppCode{}.

The following terminology and observations are useful for a further appreciation of the interface in \cref{alg:ur}.
If we add a literal $\l$ (i.e., unit clause) to a CNF, we say that we have \textit{decided} $\l$. If we later remove $\l$ from the CNF, we say that we have \textit{undecided} $\l$. In the context of SAT solving, we will make a sequence of decisions $\l_1,\ldots,\l_k$ and stop only when unit resolution (run after each decision) finds a contradiction. In that case, we will undecide some of these decisions in reverse chronological order. \cref{alg:ur} ensures that the CNF state $\ST$ is restored properly when we undecide. 
Furthermore, 
if unit resolution infers a literal $\l$, we say that literal $\l$ has been \textit{derived.} 
The states of each variable $X$ are initially \textit{active.}
If we decide or derive an $X$-literal $\l,$ then any active state in $\lnot{\l}$
will be \textit{pruned} and, hence, stop being active.
A literal $\l$ is \textit{implied} if all states in $\lnot{\l}$ are pruned.
A literal $\l$ is \textit{falsified} if all states in $\l$ are pruned.
Even though multiple (and many) literals may be derived for a particular variable during unit resolution, undoing these derivations can be done in constant 
time (see Line~\ref{ln:V-lstack} of \assert\ and Lines~\ref{ln:undo-V-prune1}~\&~\ref{ln:undo-V-prune2} of \undecide). This is in contrast to what may happen when running a Boolean SAT solver on a binarized CNF.

\cref{alg:ur} does some bookkeeping that is not needed for unit resolution but plays a key role in other ingredients of the SAT solver. We will discuss this bookkeeping next.

\begin{algorithm}[ph]
\caption{The Unit Resolution Interface (see \refAppCode{} for missing
function definitions)}\label{alg:ur}
\small
\begin{algorithmic}[1]
\Function{\decide}{$\langle V,I\rangle,\ST$}\label{alg:decide}
\Comment{variable $V$ and its states $I$ (integer)}
\State{$\ST.\dlevel \gets \ST.\dlevel + 1$} \Comment{advance decision level; $\dlevel$ initially set to $0$}
\State{$\ST.\ptime \gets 0$}
\Comment{time at which a literal is derived at this decision level}
\State{$\ST.\pvars \gets []$}
\Comment{properties at $\dlevel-1$ for variables pruned at $\dlevel$}
\State{$\success \gets \assert(\langle V,I\rangle,\nil,\ST)$}
\Comment{no reason (\nil) for this decision}
\State{push $\ST.\ptime$ to $\ST.\tstack$}
\Comment{needed to restore pruning time upon backtracking}
\State{push $\ST.\pvars$ to $\ST.\vstack$} 
\Comment{needed to restore variable properties upon backtracking}
\State{\Return{$\success$}}
\Comment{if \false, $\ST.\cclause$ is the conflict clause}
\EndFunction
\end{algorithmic}
% \vspace{.5mm}
\begin{algorithmic}[1]
\Procedure{\undecide}{$\ST$}\label{alg:undecide}
\Comment{undo the last decision}
  \State{pop $\ST.\tstack$}
  \State{$A \gets$ pop $\ST.\vstack$}
  \Comment{active states at $\dlevel-1$ for variables pruned at $\dlevel$}
  \For{$\langle V,I \rangle \in A$} 
     \Comment{variable $V$ and its active states $I$ (integer)}
    \State{$V.\astates \gets I$} \label{ln:undo-V-prune1}
    \Comment{restore active states of variable $V$ at $\dlevel-1$}
    \State{pop $V.\lstack$} \label{ln:undo-V-prune2}
    \Comment{forget that variable $V$ was pruned at $\dlevel$}
  \EndFor
  \State{$\ST.\dlevel \gets \ST.\dlevel - 1$}
  \Comment{decrement the decision level}
\EndProcedure
\end{algorithmic}
% \vspace{.5mm}
\begin{algorithmic}[1]
\Function{\assert}{$\lpair{V}{I},R,\ST$}
\Comment{asserted literal is neither implied nor falsified}
  \State{$Q \gets [(\lpair{V}{I},R)]$}
  \Comment{$R$ is the reason for the assertion (\nil\ or clause)}
  \While{$Q$ is not empty} \Comment{propagating asserted literals}
    \State{$\langle V,I\rangle,R \gets$ dequeue $Q$} \Comment{the states $I$ of variable $V$ are asserted}
    \State $P \gets V.\astates \setminus I$ \Comment{states $P$ need to be pruned}
    \For{$\vs \in P$} \Comment{see \refAppCode{} for efficiently iterating over states}
      \State{$\vs.\plevel \gets \ST.\dlevel$} 
      \label{ln:record-plevel}
      \Comment{decision level at which state $\vs$ was pruned}
      \State{$\vs.\ptime \gets \ST.\ptime$} 
      \label{ln:record-ptime}
      \Comment{time at which state $\vs$ was pruned}
      \State{$\vs.\reason \gets R$} 
      \label{ln:record-reason}
      \Comment{if $R=\nil,$ state $\vs$ was pruned by a decision; otherwise, by clause $R$}
    \EndFor{}
    \State{increment $\ST.\ptime$}
    \label{ln:inc-ptime}
    \Comment{incremented when a literal is derived}
    \If{$\ST.\dlevel \neq V.\lstack.\tops$ $\And \ST.\dlevel \neq 0$ } \Comment{$V$ not yet touched at $\ST.\dlevel$}
        \State{push $\ST.\dlevel$ to $V.\lstack$} 
        \label{ln:V-lstack}
        \Comment{record that $V$ is losing states at $\ST.\dlevel$}
        % \State{add $\langle V, V.\astates \rangle$ to $A$}
        \State{add $\langle V, V.\astates \rangle$ to $\ST.\pvars$}
        \Comment{save $V$'s active states at $\ST.\dlevel-1$}
    \EndIf{}
    \State{$V.\astates \gets V.\astates \cap I$}
    \For{$\vs \in P$} \Comment{see \refAppCode{} for efficiently iterating over states}
      \For{$\l_1 \in \vs.\watchedby$} \Comment{iterate over literals watching the pruned state $\vs$}
        \If{not $\falsified(\l_1)$} \Comment{$\l_1$ has active states}
            \State{$\vs' \gets \astate(\l_1)$} \Comment{$\l_1$ finds another active state to watch}
            \State{add $\l_1$ to $\vs'.\watchedby$; remove $\l_1$ from $\vs.\watchedby$}
        \Else{}  \Comment{$\l_1$ is falsified (the only possible case if $\l_1.\var$ is Boolean)}
            \For{each clause $\beta$ in $\l_1.\watchedby$} \Comment{iterate over clauses watching $\l_1$} \label{line:ur:iterate-l-watchedby}
              \State{$\l_2 \gets $ other literal being watched by clause $\beta$}
              \State{$\l_3 \gets$ literal in $\beta.\literals$ s.t. $\l_3 \neq \l_2$ and $\falsified(\l_3)=\false$}
              \If{literal $\l_3$ exists} 
                \If{$\l_3.\watchedby$ is empty} \label{line:ur:is-first-watching-clause} \Comment{$\beta$ is the first clause to watch $\l_3$}
                    \State{$\vs_3 \gets \astate(\l_3)$}
                    \Comment{some active state of $\l_3$}
                    \State{add $\l_3$ to $\vs_3.\watchedby$} \Comment{$\l_3$ now watches one of its active states}
                \EndIf{}
                \State{add $\beta$ to $\l_3.\watchedby$; remove $\beta$ from $\l_1.\watchedby$} \label{line:ur:update-l-watchedby}
               \Else \Comment{$\l_2$ is the only potentially active literal}
                \If{$\implied(\l_2)$}
                    {do nothing} \Comment{$\beta$ is subsumed}
                \ElsIf{$\falsified(\l_2)$}
                    \Comment{all literals of clause $\beta$ have been falsified}
                    \State{$\ST.\cclause \gets \beta$} \label{ln:save-cclause} \Comment{save conflict clause $\beta$}
                    \State{\Return{\false}} \Comment{contradiction found}
                \Else{} \Comment{$\beta$ derived literal $\l_2$}
                    \State{add $(\langle\l_2.\var,\l_2.\states\rangle,\beta)$ to $Q$}
                    \label{ln:ur-derived-l}
                \EndIf{}
              \EndIf{}
            \EndFor{} % end loop for l.watched_by
            \If{$\l_1.\watchedby$ is empty} \Comment{$\l_1$ no longer watched by any clause}
                \State{remove $\l_1$ from $\vs.\watchedby$} \Comment{$\l_1$ does not need to watch any state}\label{line:ur:l-stop-watching-s}
            \EndIf{}
        \EndIf{}
      \EndFor{} % end loop for s.watched_by
    \EndFor{} % end loop for s in P
  \EndWhile{}
\EndFunction{}
\end{algorithmic}
\end{algorithm}

\subsection{Keeping Track of the Time and Reason for Falsification}

A central notion in our proposed solver is that of \textit{falsification,}
which applies to states, literals and clauses. When a state $\vs$ of some variable $X$
is known to be impossible, we say that the state has been falsified or pruned (we use the latter term more often). When every state in a literal $\l$ has been pruned, we say that the literal has been falsified. 
When every literal in a clause has been falsified, we say the clause has been falsified (or became empty).
Literal falsification is the key activity underlying
unit resolution: when literals are falsified, clauses become shorter, leading to unit
clauses (literal derivations) or empty clauses (contradictions). As such, we need to efficiently detect when a literal has been falsified which we addressed earlier (using watched states). 
We also need to keep track of the reason for a falsification and the time at which it took place as this information will play a key role later in backtracking and clause learning when a contradiction is found (discussed in \cref{sec:cdcl}).

There are two possible reasons for pruning a state $\vs$ of some variable $X.$
The first is that we decided an $X$-literal $\l$ by calling \decide($\l,\ST$) where $\vs \in \lnot{\l}.$
The second is that we derived such a literal $\l$ because a clause $\alpha$ became unit (Line~\ref{ln:ur-derived-l} of \assert). 
The reason is stored in $\vs.\reason$ as $\nil$ in the first case,
and as clause $\alpha$ in the second case (Line~\ref{ln:record-reason} of \assert).

Falsification time is handled as follows. 
For a state $\vs$, falsification time is a pair $\langle t_1,t_2 \rangle$ where 
$t_1$ is the decision level at which state $\vs$ is pruned ($\vs.\plevel$ on Line~\ref{ln:record-plevel} of \assert) and
$t_2$ is the time at that level when the state was pruned ($\vs.\ptime$ on Line~\ref{ln:record-ptime} of \assert) ---  $\ST.\ptime$ 
is incremented when a literal is derived at that level (Line~\ref{ln:inc-ptime} of \assert).
The falsification time of a literal $\l$
is the falsification time of the state $\vs$ in $\l$ that was pruned last. 
Such a state $\vs$ is maintained only for certain literals during clause learning (Section~\ref{sec:clearn}).
Finally, the falsification time of a clause $\beta$ is 
the falsification time of the state $\vs$ in $\beta$ that was pruned last. 
We do not maintain this but will use it when making arguments
in Section~\ref{sec:clearn}.

The discrete SAT solvers discussed in~\cite{iccad/CAMA2003,cp/mvsat2010} used lazy schemes that differ from what we 
proposed above. In~\cite{iccad/CAMA2003}, two literals of each clause are watched, and when a variable loses active states, all watched literals of that variable are visited.
In~\cite{cp/mvsat2010}, each clause watches two of its states, which differs from our method. We actually started with this method before switching to the more refined and efficient method of each clause watching two literals, and each literal (that is being watched) watching one of its states.

\section{CDCL SAT Solver for Discrete CNFs}\label{sec:cdcl}

CDCL solvers learn a clause each time unit resolution discovers a contradiction~\cite{Grasp,Chaff,faia/DarwicheP21,understandCDCL}.
A learned clause is implied by the CNF, and solvers usually seek empowering learned clauses which make unit resolution more complete~\cite{aaai/PipatsrisawatD08a}.
These solvers use empowering clauses to combine search and inference as follows~\cite{faia/DarwicheP21}.
In the search part, they heuristically set variables to values (i.e., make decisions) with the hope of identifying a satisfying variable setting.
During this search process, unit resolution may discover that the current variable setting is inconsistent with the CNF. 
Inference is invoked at this point to learn an empowering clause based on the unit resolution proof of inconsistency. 
The search process is then resumed but only after undoing some of the decisions and adding the learned clause to the CNF. The number of undone decisions is determined by the learned clause as we discuss later. 
If the CNF is unsatisfiable, unit resolution will eventually become complete 
enough to establish a resolution proof of unsatisfiability without any variable settings (i.e., decisions).
%The above search-infer technique, which transformed SAT solving, is based on the ability to learn a particular type of empowering clauses called asserting clauses.

In principle, one can learn any clause as long as it
is implied by the given CNF. However,
SAT solvers usually learn \textit{asserting clauses}
which are a subset of \textit{conflict clauses}
which are relatively \textit{easy} to learn. 
Asserting clauses
are also a subset of \textit{empowering clauses}
which are \textit{useful} to learn. 
We will explain these types
of clauses next, starting with conflict clauses.

Suppose we have applied a sequence of successful decisions $\l_1, \ldots, \l_{n-1}$ to a CNF
$\Delta$ using the \decide\ function of~\cref{alg:ur},
followed by a decision $\l_{n}$ that failed.
At this point, a clause $\alpha$ is said to be a \textit{conflict clause} if it 
satisfies two conditions:
(1)~unit resolution finds $\Delta \land \lnot \alpha$ inconsistent (so $\alpha$ is implied by $\Delta$ and can be learned) and
(2)~for every literal $\l$ of $\alpha$, we can derive $\lnot{\l}$
from $\Delta \land \l_1 \ldots \land \l_{n}$ using
unit resolution~\cite{iccad/ZhangMMM01,aaai/PipatsrisawatD08a}.
A key property of a conflict clause
is that for at least one of its literals $\l$,
unit resolution cannot derive $\lnot{\l}$ from $\Delta \land \l_1 \ldots \land \l_{n}$ without involving the last decision $\l_n$ in the derivation; 
otherwise, one of the earlier decisions must have failed (by the definition of a conflict clause).

An \textit{asserting clause} $\alpha$ is a conflict clause with \textit{exactly one} literal $\l$ for which $\lnot{\l}$ cannot be derived from $\Delta \land \l_1 \ldots \land \l_{n}$ without involving the last decision $\l_n$.
This literal $\l$ is called the \textit{asserting literal} of $\alpha$. 
Moreover, the \textit{assertion level} of $\alpha$ is defined as the lowest decision level $m$ at which unit resolution can derive $\lnot{\l'}$ for all
literals $\l' \neq \l$ of clause $\alpha$.\footnote{If an asserting clause is unit, its assertion level is $0$ (no decisions are made at level $0$).}
If we undo decisions $\l_{m+1}, \ldots, \l_n$ (that is, backtrack to the assertion 
level $m$), the asserting clause $\alpha$ will become a unit clause and, therefore, 
imply the asserting literal $\l$.
An asserting clause $\alpha$ is \textit{empowering}~\cite{aaai/PipatsrisawatD08a}: it makes unit resolution more complete.
In particular,
if we learn $\alpha$ (i.e., add it to the CNF) and backtrack to the assertion level, unit resolution will derive literals that it could not have derived without this learned clause.

\subsection{Learning Discrete, Asserting Clauses}\label{sec:clearn}

We will now discuss how one can learn a discrete, asserting clause when a contradiction
is detected by unit resolution (as implemented in \cref{alg:ur}). This will constitute the second major building block for our upcoming SAT solver for discrete CNFs.
Our approach is based on lifting the definitions we presented earlier for Boolean conflict clauses and Boolean asserting clauses to a discrete setting 
(which is different from the approaches in~\cite{iccad/CAMA2003,cp/mvsat2010}).

We first note that the \textit{conflict level} is the decision level at which
a contradiction is discovered by unit resolution. This is also the level at which the clause learning process is applied, immediately after the contradiction is found.
Consider Line~\ref{ln:save-cclause} of \assert\ in \cref{alg:ur}. At this point, unit resolution has falsified every literal of clause $\beta$ so the clause became empty, we have a contradiction, and $\ST.\dlevel$ is now the conflict level. 
Moreover, clause $\beta$ is a conflict clause according to the definition we gave earlier so it is stored
in the CNF state at $\ST.\cclause.$ This clause $\beta$ is not necessarily asserting and already appears in the CNF. Our goal then is to learn a new conflict clause based on $\beta$ that is also asserting.

As in the Boolean case, proposed initially in~\cite{minisat_2004}, we will do this by iteratively revising the conflict clause using resolution until it becomes asserting: 
\begin{enumerate}
    \item $\beta=$ clause that became empty during unit resolution
    \item while more than one literal in $\beta$ has been falsified at the conflict level:
    \begin{enumerate}
        \item $\vs =$ state that was last pruned in clause $\beta$ (state $\vs$
        appears in some literal of $\beta$)
        \item $\alpha \gets \vs.\reason$ ($\vs \in \lnot{\l}$ where literal $\l$ was derived by unit resolution using clause $\alpha$)
        \item $V \gets$ variable of literal $\l$
        \item $\beta \gets$ the resolvent of clauses $\alpha$ and $\beta$ 
        on variable $V$ 
    \end{enumerate}
    \item return $\beta$
\end{enumerate}
We need to take particular measures though when implementing the above procedure due to the issue we hinted at earlier: a discrete variable has an exponential number of literals, and the learning process may create a very significant number of literals that are not part of the input CNF and that will not appear in the returned learned clause.
Due to this, our clause learning algorithm does not explicitly construct a 
conflict clause during its iterations. Instead, it 
uses $V.\cstates$ to store the states of variable $V$ that appear in the (implicit) conflict clause. Moreover, it uses $V.\lastcstate$ to store the state 
in $V.\cstates$ that was pruned last by unit resolution.
If $V.\cstates = \emptyset$, then variable $V$ does not 
appear in the conflict clause and $V.\lastcstate = \nil$. 
When the learning process terminates, we can explicitly construct the learned clause by iterating over variables $V$ with $V.\cstates \neq \emptyset,$ constructing a literal for $\langle V, V.\cstates \rangle$ and adding it to the learned clause. 
The clause learning algorithm uses $\ST.\nasserting$ to store the number of literals in the conflict clause that have been falsified at the conflict level (i.e., literals
with states pruned at the conflict level).
When $\ST.\nasserting$ is $1$, the conflict clause will be asserting
and the learning process will terminate.

\begin{algorithm}[h]
\caption{Clause Learning (see \refAppCode{} for missing
function definitions)}\label{alg:aclause-new}
\small
\begin{algorithmic}[1]
\Require{$\ST.\dlevel > 0;$
$\ST.\nasserting = 0;$   
for every variable $V$, $V.\cstates = \emptyset;$ $V.\lastcstate = \nil$ }
\Ensure{$V.\cstates$ holds the set of states of variable $V$ that appear in the conflict clause}
\Function{\acl}{$\beta,\ST$}
\Comment{all literals of clause $\beta$ are falsified}
\For{$\l  \in \beta.\literals$} \Comment{add all states in $\beta$ to the conflict clause}
    \State{$\analyzel(\l,\ST)$}\label{line:call-analyze1} 
    \Comment{adding states of literal $\l$ to conflict clause}
\EndFor{}

\While{$\ST.\nasserting > 1$}\Comment{check if we have an asserting clause}
    \State{$V \gets$ variable with the latest $V.\lastcstate$}\label{line:last-pruned-state}
    \Comment{$V$ appears in current conflict clause}
    \State{$\alpha \gets V.\lastcstate.\reason$} \Comment{$\alpha$ derived a $V$-literal that pruned $V.\lastcstate$}
    \For{$\l \in \alpha.\literals$} \Comment{$\alpha$ cannot be \nil\ here}
        \If{$\l$ is a $V$-literal}
            \State{$V.\cstates \gets V.\cstates \cap \l.\states$}\Comment{removing states from the conflict clause}\label{line:cl-remove-states}
            \State{$V.\lastcstate \gets \lastpstate(V,V.\cstates)$ }
            \Comment{$V.\cstates$ are pruned}
            \If{$V.\lastcstate.\plevel \neq \ST.\dlevel$}
            \Comment{fewer literals falsified at conflict level}
                \State{decrement $\ST.\nasserting$} 
                \Comment{no more states in $V.\cstates$ pruned at conflict level}
            \EndIf{}
        \Else{}
            \State{$\analyzel(\l,\ST)$}\label{line:call-analyze2}
            \Comment{adding states of literal $\l$ to conflict clause}
        \EndIf{}
    \EndFor{} 
\EndWhile{}
\State{// $L$ are literals of asserting clause; $\l \in L$ is asserting literal}
\State{// $m$ is assertion level; $\l_m \in L$ is last literal falsified at assertion level}
\State{\Return{$(L,\l,m,\l_m)$}} 
\Comment{returned values are described above and can be computed easily}
\EndFunction
\end{algorithmic}   

\vspace{.5mm}
\begin{algorithmic}[1]\label{alg:analyzel}
\Function{\analyzel}{$\l,\ST$}
\Comment{add states of literal $\l$ to the conflict clause}
\State{$V,I \gets \l.\var, \l.\states$ }
\Comment{\analyzel{} will add states of $I$ to $V.\cstates$}
\State{$\vs \gets V.\lastcstate$}
\Comment{state of variable $V$ that was pruned last (possibly $\nil$)}
\State{\textit{atC} $\gets s\!\neq\!\nil\; \&\; s.\plevel\!\!=\!\!\ST.\dlevel$}
\Comment{$V$-literal of conflict clause falsified at conflict level}
\State{$S \gets I \setminus V.\cstates$} \Comment{states of literal $\l$ not already in conflict clause}
\For{$\vs \in S$} \Comment{see \refAppCode{} for efficiently iterating over the states of a literal}
    \State{add $\vs$ to $V.\cstates$}
    \If{$\vs$ is $\nil$ or \pafter($\vs, V.\lastcstate$)} 
        \State{$V.\lastcstate \gets s$} \Comment{state $\vs$ is the last falsified state in $V.\cstates$}
    \EndIf{}
\EndFor{}
\State{$\updatescore (V, S, \ST)$} \Comment{needed for decision heuristic}
\If{not \textit{atC} and $V.\lastcstate.\plevel\!\!=\!\!\ST.\dlevel$}
    \State{increment $\ST.\nasserting$}
    \Comment{$V$-literal in conflict clause is now falsified at conflict level}
\EndIf{}
\EndFunction
\end{algorithmic}

\end{algorithm}

\cref{alg:aclause-new} contains the pseudocode for our clause learning method,
where \acl\ is the main function.
The code views the conflict clause (whether asserting or not) as a collection $C$ of states (from different variables) such that pruning every state in $C$ would lead to a contradiction. This clearly holds for the initial conflict clause discovered by \cref{alg:ur}, which is initially passed to \acl\ to start
the learning process. The algorithm keeps iterating as long as the conflict clause has more than one literal falsified at the conflict level ($\ST.\nasserting > 1$). In each iteration, the algorithm identifies
the state $\vs$ in collection $C$ that was pruned last by unit resolution,\footnote{The last pruned state can be found efficiently (e.g., with a heap) by getting the variable $V$ with the latest pruned state $V.\lastcstate$ (Line~\ref{line:last-pruned-state} of \acl).} 
then resolves the current conflict clause with 
clause $\alpha=\vs.\reason$ over the variable of state $\vs.$ 
This resolution step can be viewed as removing and adding states into the 
conflict clause $C$ (removal on Line~\ref{line:cl-remove-states}, and 
addition by calling \analyzel\ on 
Lines~\ref{line:call-analyze1} and~\ref{line:call-analyze2}).

To see that this learning algorithm will terminate, 
let reason $\alpha = \l \lor \alpha_0,$
conflict $\beta  = \l' \lor \beta_0,$ $\vs \in \lnot{\l}$ and $\vs \in \l'.$
Resolving $\alpha$ and $\beta$ over the variable of state $\vs$ yields
the new conflict clause $(\l \land \l') \lor \alpha_0 \lor \beta_0$.
Some states of the conflict clause $\beta$ have been lost: those in literal $\l'$ but
not in literal $\l.$ Some new states have been added to the conflict clause:
those in $\alpha_0$ but not in $\beta_0.$
By the semantics of reasons, 
the added states are guaranteed to be falsified earlier than state $\vs,$
so function \acl\ is guaranteed to terminate. In particular, while $\ST.\nasserting$
may go up or down during the iterations of \acl, the falsification time of the conflict clause decreases monotonically.

The original method for constructing asserting clauses in the Boolean setting was based
on the notion of an \textit{implication graph}~\cite{Grasp,Chaff}, which captures a trace of derived literals by unit resolution up to a contradiction. 
Certain cuts in the implication graph correspond to asserting clauses, and
some of these cuts produce more favorable clauses, called first-UIP clauses. 
The method based on resolution employed by \cref{alg:aclause-new}
produces first-UIP asserting clauses, with a complexity that is linear in the
size of reason clauses. 
A clause learning algorithm was proposed in~\cite{iccad/CAMA2003}, which produced
weak clauses as also observed later in~\cite{cp/mvsat2010} who provided a strengthening but with a costly time complexity that is lower bounded by a 
polynomial (the implication graph in~\cite{cp/mvsat2010} had pruned states as nodes instead of literals).

\subsection{Decision Heuristic}

The notion of a \textit{decision} is central to SAT solving as it corresponds
to selecting a portion of the search space to focus on by restricting the states of a variable. In a Boolean setting, the only way to restrict the value of a variable is by fixing its state to either \true\ or \false\ (i.e., assert a literal). In a discrete
setting, the number of restrictions is exponential in the cardinality of a variable,
each of which also corresponds to asserting a literal.

Decision heuristics usually pick a variable first and then select one of its literals
to assert (i.e., decide).\footnote{For example, one heuristic is to prefer variables 
with fewer active states, called the ``fail-first'' principle~\cite{haralick1980failfirst} in the constraint satisfaction literature.}
The decision heuristic in \dsat{} combines VSIDS-style scoring in Boolean SAT solvers~\cite{Chaff} with information about the active states of variables.\footnote{In Boolean CDCL solvers, variable state independent decaying sum (VSIDS)~\cite{Chaff} has been a very influential heuristic for variable ordering.
An activity score is maintained for each variable, and it is bumped when the variable is in the learned clause; in practice, variables seen during conflict resolution are often bumped as well.
The unassigned variable with the highest activity score is chosen for the next decision.
To favor the recently seen variables, all scores are halved periodically; alternatively, many solvers increase the magnitude of the bump after every conflict.
The phase of the decision variable is often selected using phase saving~\cite{sat/PipatsrisawatD07}, which chooses the variable's last assigned value (phase).} 
An activity score is maintained for each variable and state. 
For states that are added into the conflict clause by \analyzel\ of \cref{alg:aclause-new}, we bump their scores and the score of their variable. 
A literal is selected by first choosing the variable with the highest  $\frac{V.\score}{|V.\astates|}$. 
A subset of the variable's states are chosen by continuously picking the state with the top score, until the sum of the scores of the picked states reaches a certain percentage (we used 30\%) of the sum of the scores of all active states.\footnote{We found that extreme values for the threshold generally do not work well. 
In particular, choosing very few states often leads to more conflicts.
Moreover, choosing most of the states (thus pruning very few) typically does not perform well either, for the following reason.
Suppose a decision only prunes one state $s$, and some literal $\l$ in $\vs.\watchedby$ finds its last remaining active state $\vs'$ to watch, so $\l$ is moved to $\vs'.\watchedby$.
Then, when state $\vs'$ is pruned later, literal $\l$ would be visited again. 
Now, consider the scenario where the decision prunes both $\vs$ and $\vs'$. 
When the literal $\l$ is visited in $\vs.\watchedby$, $\l$ is falsified and would not be added to $\vs'.\watchedby$, so $\l$ is visited one fewer time.}
The resulting literal (states) is then asserted as a decision.
It is not uncommon to make \textit{simple decisions} in a discrete setting by
\textit{fixing} the state of a discrete variable instead of \textit{restricting} 
its set of states (\cite{iccad/CAMA2003} used simple decisions; \cite{cp/mvsat2010} 
did not specify its decision heuristic).
We tried this initially but found it less effective as it may force the solver to commit prematurely and create unnecessary conflicts.

\subsection{The \dsat\ Solver} 
\label{sec:dsat}

\begin{algorithm}[tb]
\caption{CDCL SAT Solver (see \refAppCode{} for missing
function definitions)}\label{alg:dsat-new}
\small

\begin{algorithmic}[1]
\Function{\dsat}{$\Sigma,\Delta$}
\State{// --$\Sigma$ is a set of pairs $\langle i, c \rangle$ where $i$ is a variable index and $c$ is its cardinality}
\State{// --$\Delta$ is a set of clauses, where each
clause is a set of pairs $\langle i,I\rangle$: $i$ is a variable index}
\State{// \ \ \ \ \ and $I$ is an integer representing a set of states (i.e., literal)}
\State $\ST \gets \cnfstate(\Sigma,\Delta)$
\Comment{holds CNF and bookkeeping data}
\For{each literal $\langle V, I \rangle$ in $\ST.\uclauses$} \label{ln:dsat-uc-s}
\Comment{treating unit clauses as asserted literals}
\If{$\assert(\langle V, I \rangle,\nil,\ST) = \false$}
\Comment{conflict without decisions}
\State{\Return{$\false$}} \label{ln:dsat-uc-e}
\Comment{unsatisfiable}
\EndIf{}
\EndFor{}
\While{\true}
\State{$V,I \gets \selectl(\ST)$ }
\Comment{decision heuristic: choose decision}
\If{$V = \nil$} \Comment{no more decisions to be made}
  \State{\Return $\true$}
  \Comment{satisfiable}
\EndIf{}
\State{$\success \gets \decide(\langle V,I \rangle,\ST)$}
\Comment{making a decision: literal $\langle V,I \rangle$ not implied, not falsified}
\While{$\success = \false$} \label{ln:dsat-conflict-s}
\Comment{construct asserting clause and add it at assertion level}
\State{multiply $\ST.\bump$ by $\ST.\bumpInc$} \Comment{decision heuristic: increase bump magnitude}
\State{$\beta \gets \ST.\cclause$}
\Comment{$\beta$ is conflict clause, $\ST.\dlevel$ cannot be $0$}
\State{$L, \l_a, m, \l_m \gets \acl(\beta,\ST)$} \label{ln:dsat-learn}
\Comment{must be called at conflict level}
\While{$\ST.\dlevel \neq m$} \label{ln:dsat-back1}
\Comment{backtrack to assertion level $m$}
\State $\undecide(\ST)$ \label{ln:dsat-back2}
\EndWhile{}
\State{$\success \gets \addclause(L,\l_a,\l_m,\ST)$}\label{ln:dsat-add-cl}
\Comment{must be called at assertion level}
\If{$\success=\false$ and $m=0$} \label{ln:dsat-unsat}
\Comment{conflict without decisions ($L\!=\![\l_a], \l_m=\nil$)}
\State{\Return{$\false$}}
\Comment{unsatisfiable}
\EndIf{}
\EndWhile{}\label{ln:dsat-conflict-e}
\EndWhile{}
\EndFunction
\end{algorithmic}

\vspace{.5mm}
\begin{algorithmic}[1]
\Function{\addclause}{$L,\l_a,\l_m,\ST$}
\Comment{all literals in $L$ are falsified except for $\l_a$}
\State{$\ST.\ptime \gets \ST.\tstack.\tops$}
\Comment{saved value used to resume unit resolution at assertion level}
\State{$\ST.\pvars \gets \ST.\vstack.\tops$}
\Comment{saved value used to resume unit resolution at assertion level}
\If{$L = [\l_a]$}
\Comment{unit asserting clause, $\l_m=\nil$, $\ST.\dlevel=0$}
% \State{\Return{$\assert(\l_a,\nil,\ST)$}}
\State{\Return{$\assert(\langle \l_a.\var, \l_a.\states \rangle,\nil,\ST)$}}
\Comment{this assertion will never be retracted}
\Else{}
\State{$\gamma \gets \getc(L,\l_a,\l_m,\true)$}
\Comment{constructs learned clause (will watch literals $\l_a,\l_m$)}
\State{add clause $\gamma$ to $\ST.\clauses$}
\State{\Return{$\assert(\langle \l_a.\var, \l_a.\states \rangle,\gamma,\ST)$}}
\Comment{may fail, leading to another asserting clause}
\EndIf{}
\EndFunction
\end{algorithmic}
\end{algorithm}

We are now ready to discuss the CDCL SAT solver for discrete CNFs,
the function \dsat\ in~\cref{alg:dsat-new}. 
The solver has the same overall structure as a Boolean CDCL solver. 
First, our implementation of unit resolution assumes that all clauses have
at least two literals so \dsat\ treats unit clauses in the input CNF as asserted literals $\l$ that permanently prune the states 
in $\lnot{\l}$ (Lines~\ref{ln:dsat-uc-s}-\ref{ln:dsat-uc-e}). If any of these assertions fail, the CNF is unsatisfiable;
otherwise, the assertions will not be retracted later. 
Second, as long as the CNF is not found to be unsatisfiable, 
\dsat\ keeps making decisions until a contradiction is discovered by unit resolution.
It then learns an asserting clause (Line~\ref{ln:dsat-learn}), backtracks to the assertion level by retracting enough decisions (Lines~\ref{ln:dsat-back1},~\ref{ln:dsat-back2}),
adds the learned clause to the CNF (Line~\ref{ln:dsat-add-cl}) 
and continues making decisions if the addition succeeds. 
The addition of a learned clause leads to deriving the asserting literal which may lead unit resolution to discover another contradiction which would lead to learning yet another asserting clause. If a contradiction is found by unit resolution without any decisions (Line~\ref{ln:dsat-unsat}), the CNF is unsatisfiable. This can only happen after adding a unit learned clause at decision level $0$ (i.e., the assertion level is $0$). 

\section{Experiments}
\label{sec:experiments}

We evaluated \dsat\ in three experimental settings: on randomly generated discrete CNFs at transition ratio, on CNFs generated from classical planning problems, and on the task of 
converting discrete NNF circuits into discrete CNFs which is 
relevant for explaining the decisions of classifiers. 
We will describe each experiment in more detail later.

We compared \dsat{} with representative approaches from two broad categories.\footnote{The experiment did not include the two prior works~\cite{iccad/CAMA2003,cp/mvsat2010} developing native discrete SAT solvers because the corresponding implementations are not publicly available.}
One is to directly binarize the discrete CNF and call Boolean SAT solvers;
for this, we used the PySAT library~\cite{SAT24/ignatievPysat},\footnote{\url{https://github.com/pysathq/pysat}}, which implements a variety of schemes (e.g., pairwise, sequential counters) to encode cardinality constraints (e.g., at-most-one) as clauses.
We used Cadical~\cite{CAV24/BiereCaDiCaL} with the sequential counters encoding and the pairwise encoding.\footnote{We tested all cardinality constraint encodings implemented in PySAT for all problems in the first two experiments, and found that sequential counters encoding overall performs the best.}
We also used PySAT's Gluecard4 solver, which is built on top of the Glucose SAT solver~\cite{AudemardGlucose2018} and integrates techniques first proposed in MiniCARD~\cite{SAT12/liffitonMinicard} to efficiently handle cardinality constraints.
The other category of approaches is to call solvers that handles problems with discrete variables.
We translated the discrete CNFs to the MiniZinc format~\cite{nethercote2007minizinc},\footnote{Each discrete clause is directly translated as a constraint in MiniZinc using set membership and disjunction operators. No additional variables or constraints are introduced.}
which is supported by a variety of CSP solvers. 
We used the PicatSAT solver in our experiments.\footnote{We also tried other top-performers in the recent MiniZinc challenges, OR-Tools CP-SAT~\cite{cpsatlp} and Choco-solver CP-SAT~\cite{prud2022choco}, but PicatSAT performed better across our experimental settings. }

We implemented \dsat{} in C++.\footnote{\dsat{} source code is available at \url{https://github.com/uclareasoning/dsat}.}
In addition to the pseudo-code detailed in this paper, we also implemented a few widely used techniques: clause removal, restarts, and learned clause minimization~\cite{SAT09/SrenssonMinimizeClause}, where
we used learned literal blocking (LBD)~\cite{IJCAI2009/AudemardPredicting} to guide clause removal and restarts~\cite{CP12/AudemardglucoseRestart}.
It is worth noting that, Kissat~\cite{SAT24/Bierekissat} (used by PicatSAT), Cadical, and Glucose are state-of-the-art Boolean SAT solvers that use a far wider range of techniques.
For example, a large set of techniques used by Cadical is listed in \cite{CAV24/BiereCaDiCaL}, including bounded variable elimination (BVE)~\cite{SAT05/EenBVE} in pre-processing, phase saving~\cite{sat/PipatsrisawatD07} in decision heuristic, chronological backtracking~\cite{SAT18/nadelChronological} (in addition to non-chronological backtracking), and vivification~\cite{ECAI08/PietteVivify, IJCAI17/LuoVivify}. These potent techniques have not been implemented in \dsat{},
so we will conduct a more comprehensive empirical evaluation
when they are implemented.
Yet, as the upcoming preliminary experiments show, \dsat{} can be significantly faster in some problems with variables of large cardinality.

\subsection{Random 3-CNFs}
\label{sec:exp:random}

We will first use generated random 3-CNFs to examine the behaviors of solvers when the cardinality of the variables increases.
The random 3-CNFs are parameterized by the number of variables $N$, the number of clauses $M$, and the variables all have the same cardinality $C$.
For a variable $X$ with cardinality $C$, an $X$-literal $\l$ is sampled by first choosing the size $|\l|$ of the literal uniformly at random from $1$ to $C-1$, and then choosing a set of $|\l|$ states of $X$ uniformly at random. 
For $C = 2, 4, 8, 16, 32, 48,$ and $64$, we generated problems with the value of $\frac{M}{N}$ at the transition ratio such that the problems are approximately 50\% satisfiable -- this creates challenging problems and also lets us compare how the solvers behave with both SAT and UNSAT instances.
To better gauge how the solvers scale across a wide range of $C$, for each cardinality, the number of variables $N$ are chosen so that all the problems have roughly the same difficulty for PicatSAT, and takes PicatSAT about 1-2 seconds to solve. 
As we will see in the result, this lets us select problem sizes that are not too trivial or intractable for the four solvers.
For each $C$, 200 problems instances are generated, and their parameters are shown in~\cref{tab:3cnfSizes}.
The experiment was run on an Intel i7-13700 processor, with a timeout of 600 seconds. 
Only one solver process was run at a time to minimize interference, and the process was pinned to a performance core.

\begin{table}[h]
    \centering
    \caption{Random 3-CNFs with discrete variables of cardinality $C=$ 2, 4, 8, 16, 32, 48, and 64. For each $C$, 200 problem instances at the transition ratio are generated; the number of variables $N$ and the number of clauses $M$ of the problems are shown. 
    For $C > 2$, we set $\frac{M}{N}$ to the transition ratios that are determined experimentally with a minimum of 500 problem instances.}
    \label{tab:3cnfSizes}
    \begin{tabular}{|c|c|c|c|}
    \small
      C & N & M & $\frac{M}{N}$\\ 
      \midrule
      2 & 240 & 1027 & 4.28\\
      4 & 125 & 972 & 8.1 \\
      8 & 70  & 833 & 11.9\\
      16 & 40 & 644 & 16.1\\
      32 & 24 & 499 & 20.8\\
      48 & 18 & 418 & 23.2\\
      64 & 15 & 384 & 25.6\\
    \end{tabular}
\end{table}

\begin{figure}
\centering
    \begin{subfigure}{\textwidth}
        \centering
        \includegraphics[width=\linewidth]{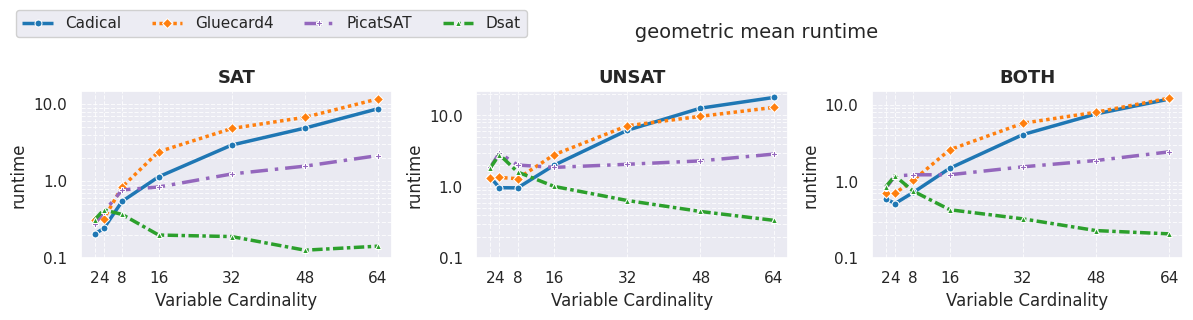}
    \end{subfigure}
    \begin{subfigure}{\textwidth}
        \centering
        \includegraphics[width=\linewidth]{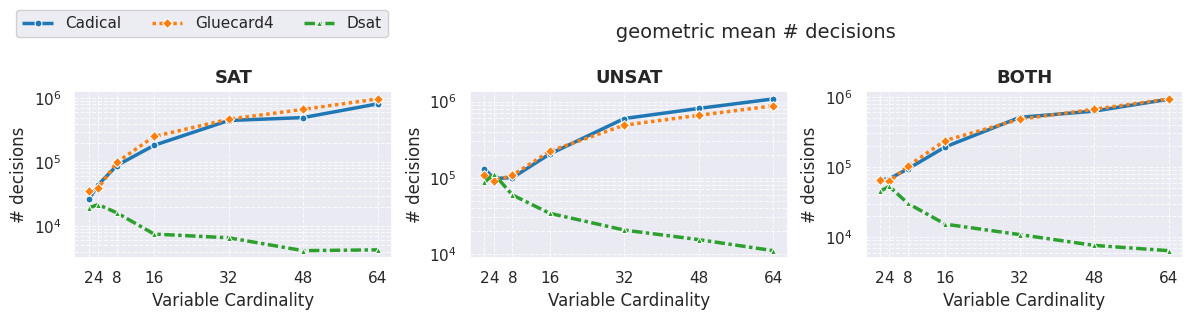}
    \end{subfigure}
    \begin{subfigure}{\textwidth}
        \centering
        \includegraphics[width=\textwidth]{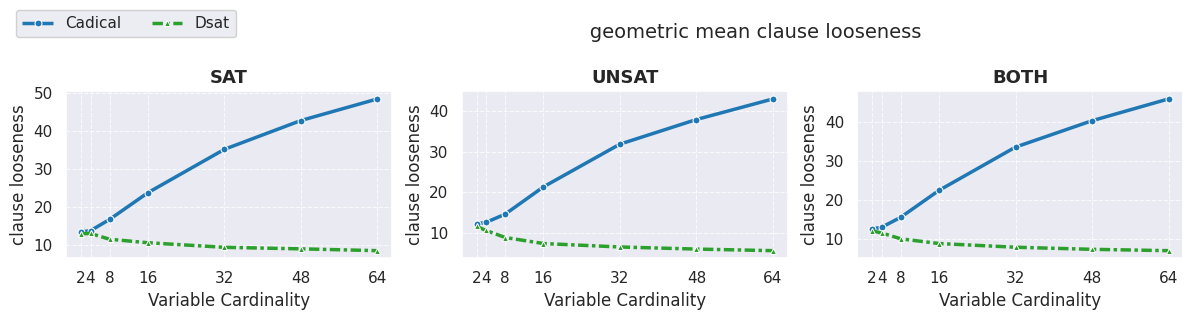}
    \end{subfigure}
    \begin{subfigure}{\textwidth}
        \centering
        \includegraphics[width=\textwidth]{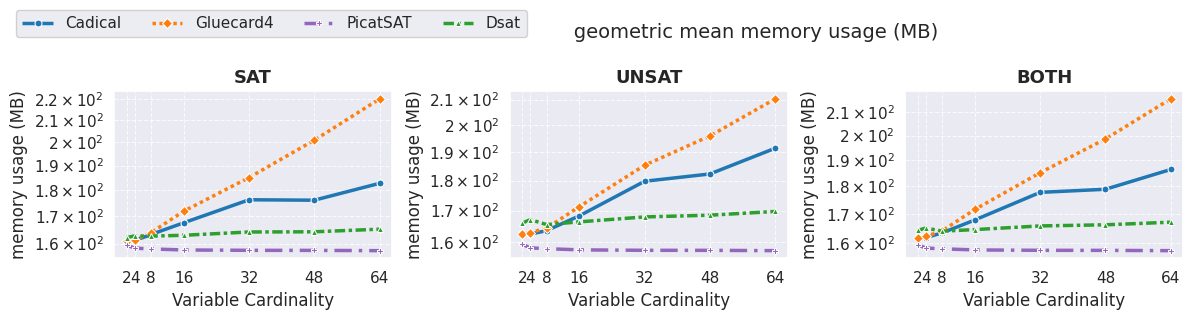}
    \end{subfigure}
\caption{
Experiment results with random 3-CNFs.
The four rows of figures from top to bottom show the geometric means of the runtime (seconds), the number of decisions, 
the strength of the learned clauses (lower looseness is stronger), and
the peak memory usage.
The x-axis in every plot is the variable cardinality $C$ in linear scale.
The top two plots (the runtime and the \#decisions) have y-axis on a log-scale.
PicatSAT is a CSP solver (called through the minizinc interface), for which we were not able to collect statistics for the \#decisions.
Only \dsat{} and Cadical provide the statistics regarding the looseness of their learned clauses.}
\label{fig:exp:random}
\end{figure}

The results are shown in~\cref{fig:exp:random}.
The top plot shows the geometric mean runtime (seconds),
and we can see that as $C$ increases, \dsat{} scales much better than the other solvers tested.
With $C = 64$, these problems take PicatSAT close to 2 seconds, and are solved by \dsat{} in about 0.1 second.
Cadical and Gluecard4 take about 10 seconds. 
For Cadical, the results with the sequential counters encoding are plotted; the results with the pairwise encoding are 10\% - 20\% slower for C = 48 and 64, but overall very similar.
The results are shown for the SAT instances and the UNSAT instances separately, as well as for all the instances combined. 
For all the solvers, the UNSAT instances generally take longer to solve than the SAT instances.
Compared to other solvers, \dsat{} seems to show a slightly larger advantage in SAT instances, but the overall trends are similar for SAT and UNSAT instances.
The plots for the number of decisions closely mirror plots for the runtime: \dsat{} makes fewer decisions compared to Gluecard4 and Cadical, and the gap widens as $C$ increases.
Furthermore, we see that \dsat{} learns stronger clauses than Cadical when variable cardinality is large, and the advantage grows with increasing $C$.
In terms of memory usage, PicatSAT is the lowest.
\dsat{} uses slightly more memory than PicatSAT, but the difference changes very little as variable cardinality increases.

\subsection{Planning}
\label{sec:exp:planning}

Since randomly generated CNFs often lack the structures in more realistic problems, we further tested the solvers using discrete CNFs generated from planning problems, which commonly involve discrete variables of large cardinality.
We will first provide some background on the ``planning as satisfiability'' (SATPLAN) formulation, and then discuss our experiment using problems in the classic blocks world domain.\footnote{We used problems from the International Planning Competition (IPC) benchmark instances collected in ~\url{https://github.com/potassco/pddl-instances}.}
It was suggested in~\cite{ecai/Kautz92_satplan} that a classical planning problem can be converted into a series of increasingly larger SAT problems.
With a given horizon $T$, a SAT problem can be created such that the problem is satisfiable iff it is possible to achieve the goal within $T$ steps.
For a planning domain with a set of state variables $\mathbf{X}$ and a set of (grounded) actions, 
at each time step $t=0, 1, 2, ..., T$, the converted SAT problem has a variable $A_t$ indicating the action taken at time $t$, and for each state variable $X \in \mathbf{X}$, there is a variable $X_t$ indicating the value of $X$ at time $t$.
The preconditions and effects of the action at step $t$ are encoded in clauses that involve the variables at time steps $t$ and $t+1$.
Note that the state variables are generally discrete, and the space of grounded actions can be large. 
For example, in the smallest blocks world problem in the experiment, with $7$ blocks, there are $15$ state variables with cardinality up to $8$, and there are $98$ grounded actions so each action variable $A_t$ has cardinality $98$; it takes on average $21$ steps to reach the goal.
In the largest problem, with $23$ blocks, there are $47$ state variables with cardinality up to $24$, and each action variable $A_t$ has cardinality $1058$; 
it takes on average $75$ steps to reach the goal.
In order to apply Boolean SAT solvers, each $X_t$ is binarized by introducing a Boolean variable for every value of $X_t$, and the same applies for each $A_t$.\footnote{This binarized problem is often directly treated in literature~\cite{sathandbook2nd/rintanen2021planning}, without describing the discrete version and the binarization step explicitly, as treated here.}

The blocks world planning instances have a number of blocks in the range $7-23$, with $2-3$ problems for each problem size, and $38$ problems in total. 
Each planning problem is first translated from the (lifted) planning domain definition language (PDDL) representation using Fast Downward~\cite{JAIR06/helmertFD}\footnote{\url{https://github.com/aibasel/downward/}} into the (grounded) SAS+~\cite{backstromSAS+1995} representation, which is then encoded into a series of $T^*$ discrete SAT problems, where $T^*$ is the shortest possible plan for the problem.
For each planning problem, we select the 5 largest SAT problems solved in SATPLAN execution: one (satisfiable) problem with horizon $T^*$ and four (un-satisfiable) problems with horizons $T^* - 1, ..., T^* - 4$; a total of 190 (non-incremental) SAT problems. 

\begin{figure}
\centering
\begin{subfigure}{0.48\textwidth}
    \centering
\includegraphics[width=\linewidth]{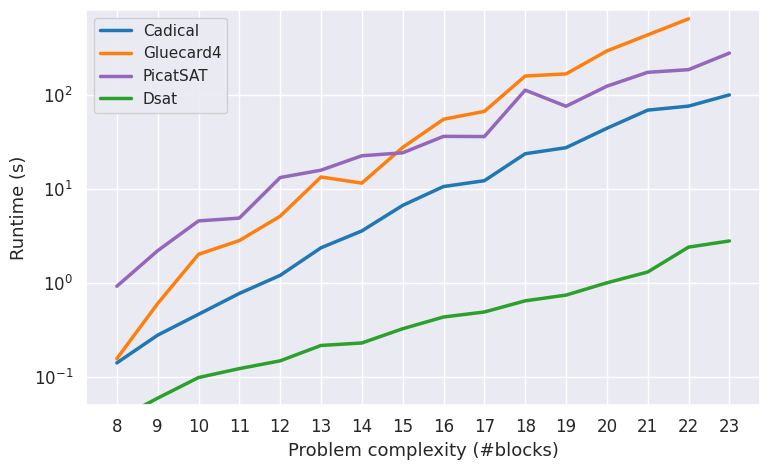}
\end{subfigure}
\begin{subfigure}{0.48\textwidth}
    \centering
    \includegraphics[width=\linewidth]{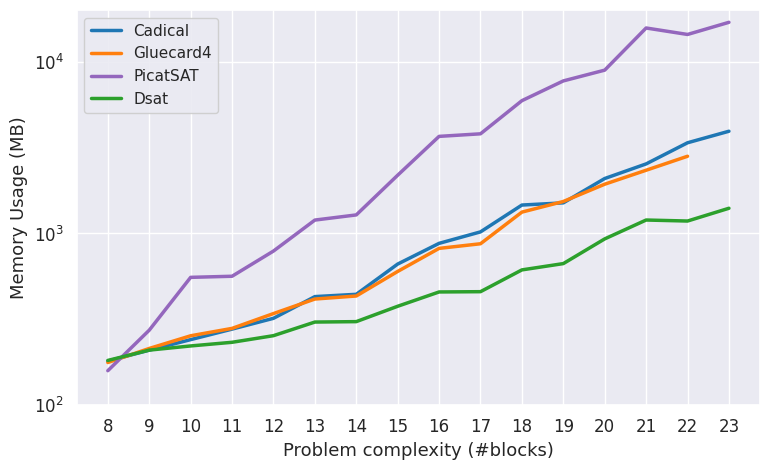}
\end{subfigure}
\caption{Comparison of \dsat, Cadical, Gluecard4, and PicatSAT on CNFs for blocks world planning problems with varying number of blocks. The geometric mean of the the runtime (left) and the memory usage (right) are plotted; both figures have their y-axis on a log-scale. The problems with 7 blocks have an average of $315$ variables and $9,653$ clauses. The problems with $23$ blocks have an average of $3,552$ variables and $663,693$ clauses.}
\label{fig:exp:blocks} 
\end{figure}

The results are shown in~\cref{fig:exp:blocks}.
Gluecard4 timed out (600s) on most instances with 23 blocks.
For Cadical, the results with the sequential counters encoding are shown; Cadical with the pairwise encoding timed out on most of the smallest instances (7-8 blocks).
In terms of both runtime and memory usage, we can see that \dsat{} performs better overall, and its advantage grows with increasing problem size.
On the hardest problems, \dsat{} is 1-2 orders of magnitude faster than the rest ($\approx\!2$ seconds for \dsat{} vs $\approx\!90$ seconds for Cadical with $23$ blocks).
It is also worth noting that the performance of SATPLAN with \dsat{} on these blocks world problems is competitive with dedicated planners like Fast Downward. 

\subsection{Converting NNFs into CNFs}

Finally, we will evaluate \dsat{} on converting discrete NNF circuits into 
discrete CNFs,
using NNF circuits which correspond to class formulas of random forest classifiers
that have been reported on in~\cite{XAI26/jiRF}. This task is relevant to
computing explanations of classifier decisions since CNF representations of
class formulas facilitate this computation.\footnote{CNFs are
or-decomposable: disjuncts do not share variables. As such, CNFs allow one to compute complete and general reasons of decisions in linear time. These reasons are the basis for computing popular explanations such as sufficient and necessary reasons; see~\cite{lics/Darwiche23} 
for a tutorial/survey.}
The conversion algorithm is classical: it traverses the NNF circuit bottom up,
computing a CNF for an AND-node by appending its children's CNFs,
and computing a CNF for an OR-node by performing a Cartesian product on
its children's CNFs. The main use of the SAT solver is to ensure that the 
CNF being computed incrementally has no subsumed clauses.
Every time some clause $\alpha = \l_1 \lor ... \lor \l_k$ is being added to the current CNF $\Delta$, we check whether $\alpha$ is subsumed by checking if $\Delta \land \lnot \l_1 ... \land \lnot \l_k$ is unsatisfiable; clause $\alpha$ is only added if it is not subsumed.
Since the conversion process may involve millions of clause subsumption checks, the algorithm (implemented in C++) uses incremental SAT solving, which is implemented in \dsat{} following the IPASIR interface~\cite{SATRace2015}.
Because incremental solving is not supported in PicatSAT and Gluecard4, we will compare \dsat{} with Cadical for this compilation task. 

\begin{figure}
\centering
\includegraphics[width=0.9\linewidth]{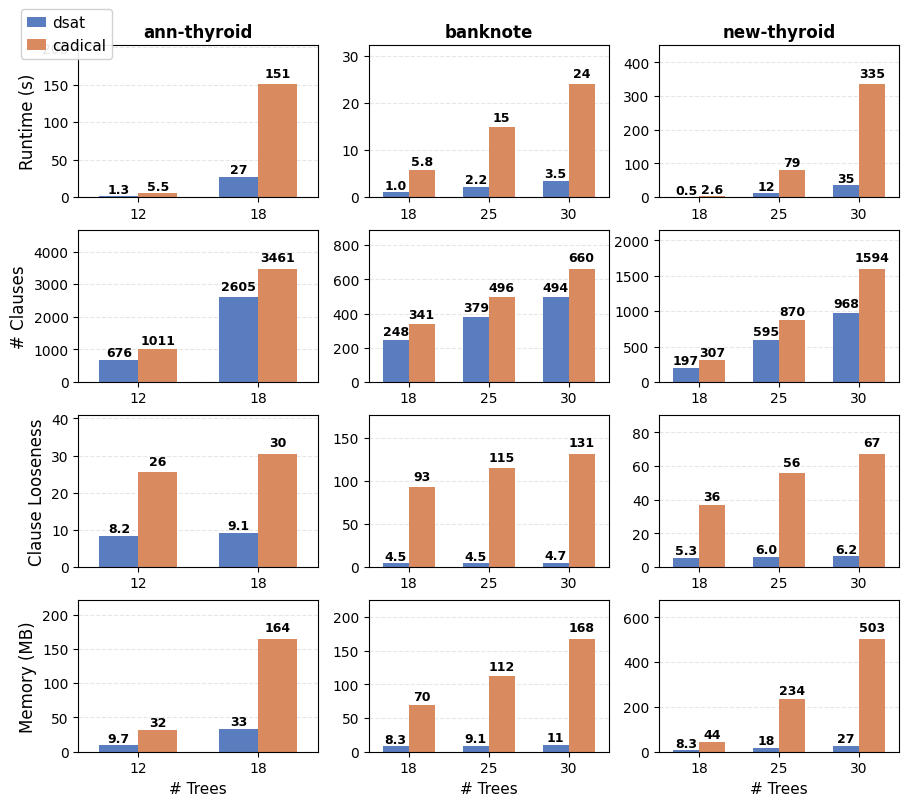}
\caption{Comparison between \dsat{} and Cadical for compiling NNF circuits into CNFs.
The four rows from top to bottom show the compilation runtime, the number of clauses in the output CNF, the average strength (lower looseness is stronger) of the clauses, and the peak memory usage.}
\label{fig:exp:compile}
\end{figure}

The experiment uses random forests trained with Weka~\cite{frank2016weka} on three datasets used in~\cite{XAI26/jiRF}, each with 2-3 output classes.
For each dataset, random forests with different sizes are trained, all with depth of 4.
Class formulas (NNFs) of the random forests are generated using~\cite{XAI26/jiRF}; there are 29 problems in total, and variables have cardinality up to 72.
\cref{fig:exp:compile} shows the experiment results; for each dataset and number of trees, the values are averaged over the dataset output classes.
Results of Cadical with the sequential counters are plotted (the pairwise encoding is about 25\% slower on the largest problems).
\dsat{} is faster overall, and shows greater advantage for larger problems; e.g., for the new-thyroid dataset with 30 trees, \dsat{} improves the compilation time by about 10x.
Furthermore, because of the native treatment of discrete variables, \dsat{} is able to better simplify the CNFs, output fewer but stronger clauses, and use less memory.

\section{Conclusion}
\label{sec:conclusion}
We presented a CDCL SAT solver for discrete CNFs,
which are CNFs in which a literal represents a set of possible values for a discrete variable. Our proposed solver is modeled around Boolean CDCL solvers with its major components being a unit resolution engine and a clause learning engine, both of which are native discrete.
Our preliminary empirical results suggest that operating on discrete variables natively during SAT solving is a promising direction, in comparison to using Boolean/hybrid SAT solvers with binarization techniques.

\bibliography{refs} %% Please use bibtex

\appendix

\newpage
\section{Data Structures and Additional Pseudocode}
\label{app:pcode}

The main objects in the \dsat\ solver 
are CNF-state, clause, variable, literal and state. The main attributes of these objects are described below. 

\vspace{3mm}
\noindent CNF-state $\ST$ attributes:
\begin{itemize}
    \item $\ST.\vcount$: Number of variables in CNF.
    \item $\ST.\var$: Array of variable objects.
    \item $\ST.\dlevel$: Decision level, initially $0$ and incremented each time a decision is made.
    \item $\ST.\uclauses$: The unit clauses in the input CNF.
    \item $\ST.\clauses$: The non-unit clauses in the input CNF.
    \item $\ST.\cclause$: Conflict clause, set by resolution when a contradiction is found and later used by the clause learning algorithm.
    \item $\ST.\ptime$: Integer that keeps track of state pruning time at 
    each decision level (for computing state falsification time). Initialized
    at each decision level, and incremented when a literal is derived at that level.
    \item $\ST.\pvars$: List of pairs $\langle V, I \rangle$ where $V$ is a variable
    and $I$ represents a set states of $V.$ Used to save active states of variables at each decision level (for rolling back CNF state).
    \item $\ST.\tstack$: Stack for saving $\ST.\ptime$ at each decision level (for rolling back CNF state).
    \item $\ST.\vstack$: Stack for saving $\ST.\pvars$ at each decision level (for rolling back CNF state).
    \item $\ST.\nasserting$: Number of variables in the conflict clause that have states pruned at the conflict level. This is used by the clause learning algorithm.
\end{itemize}

\noindent Variable $V$ attributes:
\begin{itemize}
    \item $V.\astates$: Integer representing the active states of variable $V.$
    \item $V.\lstack$: Stack that holds decision levels at which states of variable $V$ were pruned.
    \item $V.\cstates$: The states of variable $V$ in the current conflict clause. This is used by the clause learning algorithm and it implicitly means that the conflict clause has a $V$-literal with states $V.\cstates.$ If $V.\cstates = 0$, variable $V$ does not appear in the conflict clause.
    \item $V.\lastcstate$: The state in $V.\cstates$ that was pruned last.
\end{itemize}

\noindent Clause $\beta$ attributes:
\begin{itemize}
    \item $\beta.\literals$: Literals of clause $\beta.$
\end{itemize}

\noindent Literal $\l$ attributes:
\begin{itemize}
    \item $\l.\var$: The variable of literal $\l.$
    \item $\l.\states$: Integer representing the states of literal $\l$ (some active, some pruned).
    \item $\l.\watchedby$: The clauses watching literal $\l.$
\end{itemize}

\noindent State $\vs$ attributes:
\begin{itemize}
    \item $\vs.\plevel$: The decision level at which state $\vs$ was pruned.
    \item $\vs.\ptime:$ The time at which state $\vs$ was pruned (time is local to a decision level). 
    \item $\vs.\reason:$ The reason for pruning state $\vs$ (a clause or \nil).
    \item $\vs.\watchedby$: The literals watching state $\vs$ (all such literals must be watched by clauses). 
\end{itemize}

To leverage efficient bit-wise operations, a set of states is internally represented in \dsat{} as an array of integers, which we refer to as an integer for simplicity.
When iterating over the 
states represented by integer $I$ in the paper, we write ``for $\vs \in I$.'' 
In this appendix, we flesh this out to ``for $i \in \allbits(I), \vs \gets V.\state[i]$ 
which iterates over the indices of states in $I,$ then recovers the 
state object from its variable object.

\newpage

\begin{algorithm}[h]
\caption{Constructing CNF state with Watched States and Literals \label{alg:cnf-state}}
\small

\begin{algorithmic}[1]
\Function{\cnfstate}{$\Sigma,\Delta$}
\Comment{see \dsat\ for definition of $\Sigma, \Delta$}
\State{$\ST \gets $ new CNF object}
\Comment{fields not initialized: $\ptime, \pvars, \cclause$}
\State{$\ST.\vcount \gets |\Sigma|$}
\Comment{number of variables}
\State{$\ST.\var \gets $ array of size $|\Sigma|$}
\Comment{maps variable index to variable object}
\State{$\ST.\clauses \gets []$}
\Comment{list of non-unit clauses}
\State{$\ST.\uclauses \gets []$}
\Comment{list of literals corresponding to unit clauses}
\State{$\ST.\dlevel \gets 0$}
\Comment{decision level}
\State $\ST.\tstack \gets \langle 0 \rangle$
\Comment{stack with one entry for $\ptime$ (pruning time)}
\State $\ST.\vstack \gets \langle [] \rangle$
\Comment{stack with one entry for $\pvars$ (properties of variables)}
\State{$\ST.\bump \gets 1$}
\Comment{used to update scores for the decision heuristic}
\State{$\ST.\bumpInc \gets 1.05$}
\Comment{how much to scale $\ST.\bump$ at every conflict (configurable)}

\For{each $\langle i, c \rangle$ in $\Sigma$}
  \Comment{construct variable and state objects}
  \State{// $i$ is a variable index, $c$ is its cardinality}
  \State{$V \gets $ new variable object}
  \Comment{fields not initialized: none}
  \State{$\ST.\var[i] \gets V$}
  \Comment{so we can retrieve a variable object based on its index}
  \State{$V.\idx \gets i$}
  \Comment{variable index}
  \State{$V.\card \gets c$}
  \Comment{variable cardinality}
  \State{$V.\score \gets 0$}
  \Comment{variable activity score}
  \State{$V.\astates \gets 2^c-1$}
  \Comment{integer representing active states of $V$ (all active initially)}
  \State{$V.\state \gets$ array of size $c$}
  \Comment{maps state index to state object}
  \State{$V.\literal \gets$ hashmap}
  \Comment{maps integer to literal object}
  \State{$V.\lstack \gets $ empty stack}
  \Comment{used by \decide/\undecide}

  \For{$j=0$ to $c-1$}
  \Comment{construct state objects}
    \State{$\vs \gets$ new state object}
    \Comment{fields not initialized: $\plevel, \ptime, \reason$}
    \State{$V.\state[j] \gets \vs$}
    \Comment{so we can retrieve a state object based on its index}
    \State{$\vs.\idx \gets j$}
    \Comment{state indices start at $0$}
    \State{$\vs.\watchedby \gets \{\}$}
    \Comment{multiple literals may watch a state}
    \State{$\vs.\score \gets 0$}
    \Comment{state activity score}    

  \EndFor{}
\EndFor{}
\For{each clause $\beta$ in $\Delta$}
  \Comment{construct clause and literal objects}
  \State{$L \gets []$}
  \Comment{literal objects for clause}
  \For{each $\langle i, I \rangle$ in $\beta$}
    \Comment{construct literal objects}
    \State{// $i$ is a variable index, $I$ is an integer representing states}
    \State{$V \gets \ST.\var[i]$}
    \Comment{variable of literal}
    \State{$\l \gets \getl(V,I)$}
    \Comment{literal $\l$ may have already existed}
    \State{add $\l$ to $L$}
  \EndFor{}
  \If{$L=[\l]$}
    {add $\l$ to $\ST.\uclauses$}
    \Comment{$\beta$ is a unit clause}
  \Else
    \State{$C \gets \getc(L,L[0], L[1], \false)$}
    \Comment{watch two arbitrary literals in clause}
    \State{add $C$ to $\ST.\clauses$}
  \EndIf{}
\EndFor{}
\State{\Return{$\ST$}}
\Comment{CNF-state object}
\EndFunction
\end{algorithmic}
\end{algorithm}

\newpage

\begin{algorithm}[h]
\caption{Functions for Clauses, Variables, Literals, States and Integers}
\small
\vspace{.5mm}
\begin{algorithmic}[1]
\Function{\implied}{$\l$}\label{alg:implied}
\Comment{whether literal $\l$ is implied}
\State{\Return{$\l.\states \bAND \l.\var.\astates = \l.\var.\astates $}}
\Comment{bitwise operation}
\EndFunction
\end{algorithmic}

\vspace{.5mm}
\begin{algorithmic}[1]
\Function{\falsified}{$\l$}\label{alg:falsified}
\Comment{whether literal $\l$ is falsified}
\State{\Return{$\l.\states \bAND \l.\var.\astates = 0$}}
\Comment{bitwise operation}
\EndFunction
\end{algorithmic}

\vspace{.5mm}
\begin{algorithmic}[1]
\Function{\astate}{$\l$}
\Comment{returns an arbitrary active state in literal $\l$}
\State{$i \gets \onebit(\l.\states)$}
\Comment{$\l.\states \neq 0$}
\State{\Return{$\l.\var.\state[i]$}}
\EndFunction
\end{algorithmic}

\vspace{.5mm}
\begin{algorithmic}[1]
\Function{\pafter}{$\vs_1,\vs_2$}
\Comment{whether state $\vs_1$ was pruned after state $\vs_2$}
\State{\Return{$\vs_1.\plevel > \vs_2.\plevel$ or ($\vs_1.\plevel = \vs_2.\plevel$ and $\vs_1.\ptime > \vs_2.\ptime$)}}
\EndFunction
\end{algorithmic}

\vspace{.5mm}
\begin{algorithmic}[1]
\Function{\onestate}{$V$}
\Comment{whether variable has a single active state}
\State{\Return{$\singlebit(V.\astates)$}}
\Comment{bitwise operation}
\EndFunction
\end{algorithmic}

\vspace{.5mm}
\begin{algorithmic}[1]
\Function{\lastpstate}{$V,I$}
\Comment{returns last pruned state in states $I \neq 0$}
\State{$\vs \gets \nil$}
\For{$i \in \allbits(I)$}
\Comment{iterate over indices of on-bits in integer $I$}
  \State{$\vs' \gets V.\state[i]$}
  \If{$\vs=\nil$ or $\pafter(\vs',\vs)$}
    \State {$\vs \gets \vs'$}
  \EndIf{}
\EndFor{}
\State{\Return{$\vs$}}
\EndFunction
\end{algorithmic}

\vspace{.5mm}
\begin{algorithmic}[1]
\Function{\getl}{$V,I$} 
\Comment{variable $V$, integer $I$ representing some states of $V$}
\State{$\l \gets V.\literal[I]$}
\Comment{lookup literal}
\If{$\l \neq \nil$}
  {\Return{$\l$}}
\Comment{a literal with states $I$ already exists}
\EndIf{}
\State{$\l \gets$ new literal object}
\Comment{fields not initialized: none}
\State{$V.\literal[I] \gets \l$}
\Comment{cache literal $\l$ so we do not reconstruct it}
\State{$\l.\var \gets V$}
\State{$\l.\states \gets I$}
\Comment{an integer representing states of literal $\l$}
\State{$\l.\watchedby \gets \{\}$}
\Comment{a literal may be watched by multiple clauses}
\State{\Return{$\l$}}
\EndFunction
\end{algorithmic}

\vspace{.5mm}
\begin{algorithmic}[1]
\Function{\getc}{$L,\l_1,\l_2, \lclause$} 
\Comment{$L$ has $\geq 2$ literals, $\lclause$ indicates a learned clause}
\State{$C \gets$ new clause object}
\Comment{fields not initialized: none}
\State{$C.\literals \gets L$}

\For{$\l \in \{\l_1, \l_2\}$} \Comment{watching literals and states}
    \If{$\l.\watchedby$ is empty} \Comment{clause $C$ is the first to watch literal $\l$}
        \State{$j \gets \onebit{}(\l.\states)$} 
        \State{add $\l$ to $\l.\var.\state[j].\watchedby$}
        \Comment{watch arbitrary state in literal $\l$}
    \EndIf{}
    \State{add $C$ to $\l.\watchedby$}
\EndFor{}

\State{\Return{$C$}}
\EndFunction
\end{algorithmic}

\vspace{.5mm}
\begin{algorithmic}[1]
\Function{\onebit}{$I$}
\Comment{integer $I\neq 0$ (non-empty set of states)}
\State{\Return{INT($\log_2 I)$}}
\Comment{most-significant on-bit in integer $I$}
\EndFunction
\end{algorithmic}

\vspace{.5mm}
\begin{algorithmic}[1]
\Function{\allbits}{$I$}
\Comment{integer $I \neq 0$ (non-empty set of states)}
\State \Return{list of indices for on-bits in integer $I$}
\EndFunction
\end{algorithmic}

\vspace{.5mm}
\begin{algorithmic}[1]
\Function{\singlebit}{$I$}
\Comment{whether integer $I$ has a single on-bit}
\State{\Return{$(I > 0)$ and $(I \bAND (I - 1) = 0)$}}
\EndFunction
\end{algorithmic}

\end{algorithm}

\begin{algorithm}[h]
\caption{Decision Heuristic \label{alg:heuristic}}
\small

\vspace{.5mm}
\begin{algorithmic}[1]
\Function{\selectl}{$\ST$}
\State{$V^*, t^* \gets \nil, -\inf$}
\Comment{current best}
\For{each variable $V$ in $\ST$} \Comment{pick a variable to constrain with a decision}
  \If{$\onestate(V)$}
    {\bf Continue}
    \Comment{$V$ has one active state}
  \EndIf{}
  \State{$t \gets {V.\score}/{|V.\astates|}$}
  \If{$t > t^*$}
    \Comment{found a better variable}
    \State{$V^*, t^* \gets V, t$}
  \EndIf{}
\EndFor{}
\If{$V^* = \nil$}
  \Comment{every variable has one active state or is no longer constrained}
  \State{\Return{$(\nil,\nil)$}}
\EndIf{}
\State{// $\texttt{aSum}$ = sum of the scores of active states, $\texttt{sSum}$ = sum of the scores of selected states }
\State{// while $\texttt{sSum} \le 0.3 * \texttt{aSum}$, keep selecting the active state with the highest score}
\State{// $I$ represents the selected states}
\State{\Return{$\langle V^*,I\rangle$}}
\EndFunction
\end{algorithmic}

\vspace{.5mm}
\begin{algorithmic}[1]
\Function{\updatescore}{$V,I, \ST$}
\Comment{bumping variable and state scores in the VSIDS style}
\For{$i \in \allbits{}(I)$}
\Comment{states $I$ (integer) of variable $V$ are pruned}
    \State{increment $V.\states[i].\score$ by $\ST.\bump$}
\EndFor{}
\State{increment $V.\score$ by $\ST.\bump * (|I.\states|/V.\card)$} 
\Comment{$|I.\states|$ is number of states $I$}
\EndFunction
\end{algorithmic}

\end{algorithm}

\clearpage

%%%%%%%%%%%%%%%%%%%
%%%%%%%%%%%%%%%%%%%

\end{document}

%% file: figures/class_formula.tex
 \scalebox{0.75}{
        \begin{tikzpicture}[
        roundnode/.style={circle ,draw=black, thick},
        squarednode/.style={rectangle, draw=black, thick},
        ]
        %Nodes
        \node[squarednode]     (X)                              {\Large X};
        \node[roundnode]     (c1)       [below=of X, xshift = -0.8cm, yshift = 0.5cm] {\Large $c_1$};
        \node[squarednode]     (Y)       [below=of X, xshift = 0.8cm, yshift = 0.5cm] {\Large $Y$};
        \node[squarednode]       (Z1)   [below=of Y, xshift = -0.8cm, yshift = 0.5cm] {\Large $Z$};
        \node[squarednode]       (Z2)   [below=of Y, xshift = 0.8cm, yshift = 0.5cm] {\Large $Z$};
        \node[roundnode]       (c2)   [below=of Z1, xshift = -0.8cm, yshift = 0.5cm] {\Large $c_1$};
        \node[roundnode]       (c3)   [below=of Z1, xshift = 0.8cm, yshift = 0.5cm] {\Large $c_2$};
        \node[roundnode]       (c4)   [below=of Z2, xshift = 0.8cm, yshift = 0.5cm] {\Large $c_3$};
        
        %Lines
        \draw[-latex, thick] (X.240) -- node [anchor = center, xshift = -4mm, yshift = 1mm] {$x_1x_2$} (c1.north);
        \draw[-latex, thick] (X.300) -- node [anchor = center, xshift = 4mm, yshift = 1mm] {$x_3$} (Y.north);
        \draw[-latex, thick] (Y.240) -- node [anchor = center, xshift = -4mm, yshift = 1mm] {$y_1$} (Z1.north);
        \draw[-latex, thick] (Y.300) -- node [anchor = center, xshift = 4mm, yshift = 1mm] {$y_2y_3$} (Z2.north);
        \draw[-latex, thick] (Z1.240) -- node [anchor = center, xshift = -4mm, yshift = 1mm] {$z_1z_3$} (c2.north);
        \draw[-latex, thick] (Z1.300) -- node [anchor = center, xshift = -4mm] {$z_2$} (c3.north);
        \draw[-latex, thick] (Z2.240) -- node [anchor = center, xshift = 4mm] {$z_2$} (c3.north);
        \draw[-latex, thick] (Z2.300) -- node [anchor = center, xshift = 4mm, yshift = 1mm] {$z_1z_3$} (c4.north);
        \end{tikzpicture}
        }